\documentclass[review]{elsarticle}

\usepackage[margin=0.618in]{geometry}
\usepackage{graphicx,subcaption}
\usepackage{multirow}
\usepackage{hyperref}
\usepackage{enumitem}

\journal{Journal of \LaTeX\ Templates}
\graphicspath{{figs/}{images/}}

\begin{document}

\begin{frontmatter}
\title{Scale-Invariant Structure Saliency Selection for Fast Image Fusion}
\author{Yixiong Liang}
\ead{yxliang@csu.edu.cn}
\author{Yuan Mao}
\ead{yuanmao@csu.edu.cn}
\author{Jiazhi Xia}
\ead{xiajiazhi@csu.edu.cn}
\author{Yao Xiang}
\ead{yao.xiang@csu.edu.cn}
\author{Jianfeng Liu\corref{mycorrespondingauthor}}
\cortext[mycorrespondingauthor]{Corresponding author at: School of Information Science and Engineering, Central South University, Changsha 410083, China}
\ead{ljf@csu.edu.cn}
\address{School of Information Science and Engineering, Central South University, Changsha 410083, China}

\begin{abstract}
In this paper, we present a fast yet effective method for pixel-level scale-invariant image fusion in spatial domain based on the scale-space theory. Specifically, we propose a scale-invariant structure saliency selection scheme based on the difference-of-Gaussian (DoG) pyramid of images to build the weights or activity map. Due to the scale-invariant structure saliency selection, our method can keep both details of small size objects and the integrity information of large size objects in images. In addition, our method is very efficient since there are no complex operation involved and easy to be implemented and therefore can be used for fast high resolution images fusion. Experimental results demonstrate the proposed method yields competitive or even better results comparing to state-of-the-art image fusion methods both in terms of visual quality and objective evaluation metrics. Furthermore, the proposed method is very fast and can be used to fuse the high resolution images in real-time. Code is available at \url{https://github.com/yiqingmy/Fusion}.
\end{abstract}


\end{frontmatter}

\section{Introduction}
Pixel-level image fusion intends to combine different images of the same scene by mathematical techniques in order to create a single composite image that will be more comprehensive and thus, more useful for human or machine perception \cite{stathaki2011image,li2017pixel}. For instance, multi-modal image fusion \cite{du2016overview} tries to fuse images which have been acquired via different sensor modalities exhibiting diverse characteristics for a more reliable and accurate medical diagnosis. Another typical application is the multi-focus image fusion \cite{duan2018multifocus}. As the depth-of-field (DoF) of bright-field microscopy is only about 1\textasciitilde 2 micrometers, while the specimen's profile covers a much larger range and then the parts of the specimen that lie outside the object plane are blurred. The multi-focus image fusion can obtain an all-in-focus image from multiple images taken under different distance from the object to the lens of the identical view point.

A good image fusion method should contain the following properties. First, it preserves both the details of small size objects and the integrity information of large size objects in the fused image, even in the case of the size of the interested objects varying largely in the image. For example, the cervical cell images from the microscope contain both small size isolated cells and large size agglomerates, which are both useful for cervical cytology \cite{nayar2015bethesda}. Second, it should be efficient enough to handle large-scale data. For instance, it needs to process thousands of fields of view (FoV) in an acceptable time for the whole slide scanning in digital cytopathology \cite{pantanowitz2009impact}, which requires to fuse a series of high resolution images captured at each FoV in a very efficient way. Third, it does not produce obvious artifacts. Despite being studied extensively, to our best knowledge, existing fusion methods may not meet these requirements simultaneously.

\begin{figure}[t]
	\centering
	\includegraphics[width=0.6\linewidth]{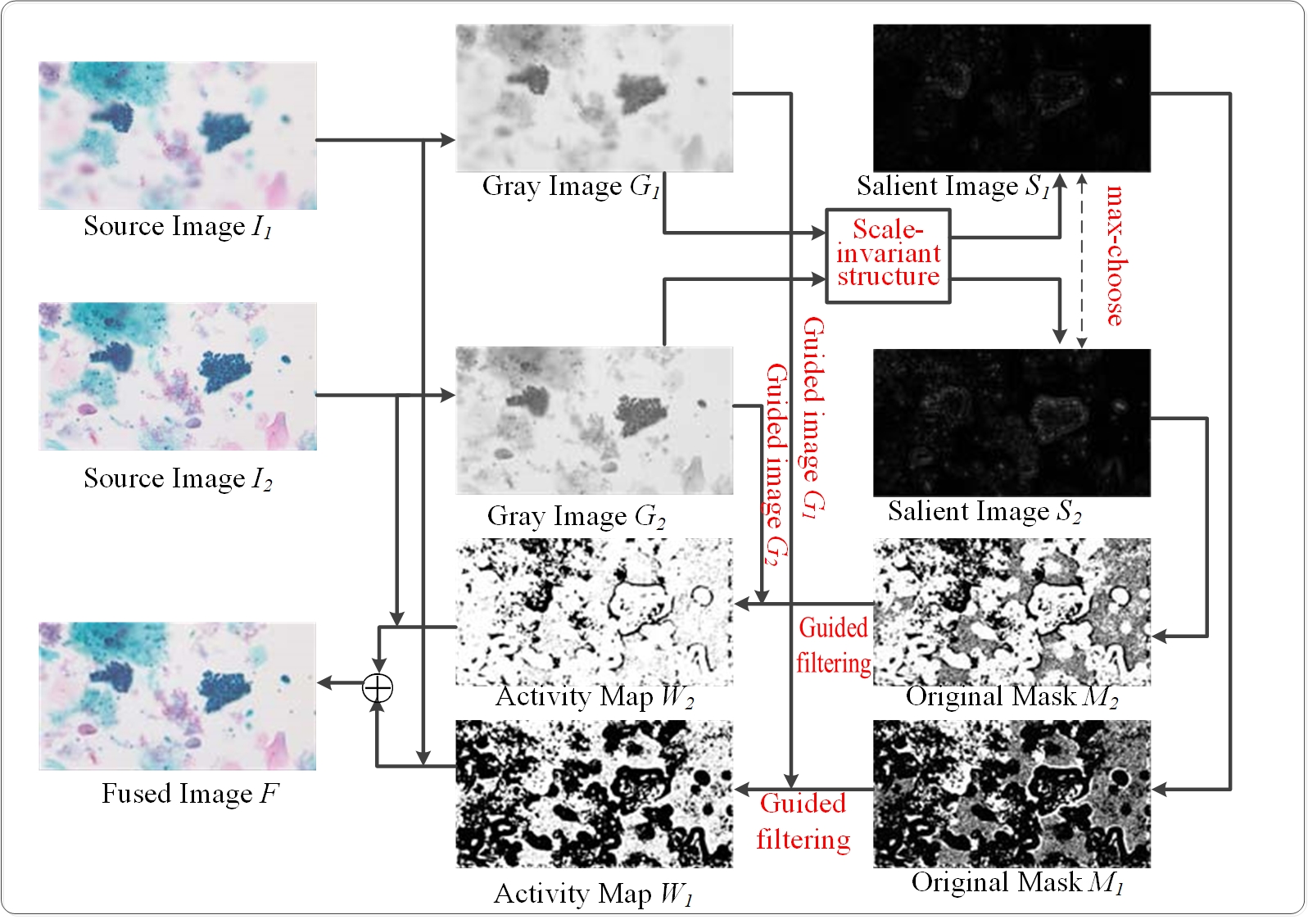}
	\caption{The proposed scheme.}
	\label{schematic}
\end{figure}

In this paper, we propose a simple yet effective image fusion method which can deal with the case where the size of the interested objects varies largely in the image, which is illustrated in Figure \ref{schematic}. It is a spatial domain method and the key idea is the scale-invariant structure saliency generation based on the difference-of-Gaussian (DoG) pyramid \cite{lowe2004distinctive}. After generating the saliency map of each image, a simple max operation is applied to them to generate the mask images, which are further refined by a single-scale guided filtering \cite{he2013guided} to exploit the spatial correlation among adjacent pixels, resulting a scale-invariant estimation of activity maps. Without complicated processing involved, the proposed method is very fast and can be used to fuse the high resolution images in real-time applications. Experimental results demonstrate that comparing to many state-of-the-art methods, the proposed method is much faster while yields competitive or even better results in terms of both visual and quantitative evaluations. Our contributions in this paper are as follows:
\begin{itemize}
\item We propose a scale-invariant structure saliency selection scheme based on the difference-of-Gaussian (DoG) pyramid of images. The resulting image fusion method can keep both details of small size objects and the integrity information of large size objects in images simultaneously.
\item Our method is very efficient, easy to be implemented and can be used for fast high resolution images fusion.
\item Comparing to many state-of-the-art methods, our method yields competitive or even better results in terms of both visual and quantitative evaluations on three datasets.
\end{itemize}

\subsection{Related works}
Various of image fusion techniques have been proposed in literatures which can be roughly classified into two categories \cite{stathaki2011image}: transform domain methods and spatial domain methods. The transform domain methods are mainly based on the ``decomposition-fusion-reconstruction'' framework, which first transforms each source image into a new domain by some tools such as multi-scale decomposition (MSD) \cite{he2018multi,liu2017structure}, or sparse coding \cite{yang2010multifocus,liu2016image,zhang2018robust} or other transformation like principal component analysis (PCA) \cite{shahdoosti2016combining}, etc., and then constructs a composite representation in the transform domain with specific fusion rules and finally applies the inverse transform on the composite representation to obtain the fused image. The spatial domain method is to take each pixel in the fused image as the weighted average of the corresponding pixels in the input images, where the weights or \emph{activity map} are often determined according to the saliency of different pixels \cite{gangapure2015steerable} and the corresponding spatial context information \cite{li2013image,chen2018robust,li2013image2,Zhang2017Boundary}. Recently there are emerging deep learning-based fusion methods \cite{liu2018deep,Liu2017Multi,du2017image} which learn the \emph{saliency map} and fusion rule based on convolutional neural network (CNN). Here we will review some related methods and explicitly distinguish them from the proposed method. Some comments are given in order:
\begin{itemize}
  \item The multi-scale fusion has the advantages of extracting and combining salient features at different scales, which is widely used in the transform domain methods. The pyramid transform and wavelet transform are the two most used categories of multiple scale decomposition schemes. For instance, the nonsubsampled contourlet transform (NSCT) \cite{zhang2009multifocus} and the dual tree complex wavelet transform (DTCWT) \cite{zhou2014multi} are used to decompose the image into a serials of subbands and then perform coefficients fusion in the transform domain, while the method of LP-SR \cite{liu2015general} uses the multi-scale Laplacian pyramid transform. Instead of performing coefficient at each scale individually, the cross-scale coefficient selection method \cite{shen2013cross} calculates an optimal set of coefficients for each scale in the transform domain. Different from these methods, our method perform multi-scale fusion directly in the spatial domain.
  \item The method of multi-scale weighted gradient fusion (MWGF) \cite{zhou2014multi} tries to combine all the important gradient information from the input images into an optimal weighted gradient image based on a two-scale scheme. Then reconstructs a fused image based on the approximately optimal weighted gradient image. However, such a simple two-scale scheme can not handle large scale variation of objects, which will result in blur or distortion in the final fused image. Moreover, the reconstruction needs to minimize the energy function which may be time-consuming. Our method searches across the entire scale-space and the reconstruction only needs to calculate the weighted average of the corresponding pixels intuitively in the input images so that it can be very efficient.
  \item The GFF-based method \cite{li2013image} also employs the guided filtering to combine pixel saliency and spatial context for image fusion. However, the pixel saliency calculated in the GFF-based method can only extract salient features at two scales. Moreover, for each source image, the GFF-based method needs to perform guided filtering twice to refine the weight maps of base layer and detail layer separately, which is a bit time-consuming. Due to the scale-invariant saliency selection, our method can deal with the case of large scale variation and is faster than GFF-based method.
  \item The CNN-based image fusion methods \cite{Liu2017Multi,du2017image} try to use CNN to learn the saliency map of each images. Specifically, the CNN models for patch similarity comparison \cite{zagoruyko2015learning} are trained using high-quality image patches and their blurred versions to encode the mapping between the source image and the corresponding saliency map \cite{Liu2017Multi}. Obviously, those CNN-based methods need to perform very complicated training. In order to deal with more than two images, the CNN-based methods try to fuse them one by one in series making the inference is very time-consuming.
\end{itemize}

\section{The Proposed Method}
\subsection{Scale-Invariant Saliency Selection}
Our scale-invariant saliency selection scheme is based on the scale space theory  \cite{lindeberg1994scale}. It has been shown that under a variety of reasonable assumptions, the only possible scale space kernel is the Gaussian function, and the scale space of an image can be produced by convolving the image with variable-scale Gaussian. Here we adopt the scheme in \cite{lowe2004distinctive} to generate the sampled scale space. The initial image is incrementally convolved with Gaussian function to produce images separated by a constant factor $k$ in scale space and each octave of scale space is divided into several layers with an integer number $s$, where $k=2^{1/s}$. Once a complete octave has been processed, the first Gaussian image of the next octave has twice the initial value of $\sigma$ and is down-sampled by taking every second pixel in each row and column of the Gaussian image in its previous octave.

We then utilize the sampled scale space of image $I(x,y)$ to generate the corresponding scale-invariant saliency map $S(x,y)$. Inspired by \cite{lowe2004distinctive}, we first define a \emph{scale-dependent saliency metric} $S_\sigma$ based on the difference of Gaussian (DoG) response which can reflect the local image structure at the current scale $\sigma$ and is robust to noise as follows
\begin{equation}\label{dog_metric}
S_{\sigma}(x,y) = g(\sigma_I) \ast |L(x, y, \sigma) - L(x, y, k\sigma)|,
\end{equation}
where  $L(x, y, \sigma) = I(x,y)\ast g(\sigma)$ is the Gaussian image, $g(\sigma) = \frac{1}{2\pi \sigma^2} \exp (-\frac{x^2 + y^2}{2 \sigma^2})$ is the Gaussian function and $\ast$ is the convolution operator. The absolute values of the DoG response are further averaged in the neighborhood of the point by smoothing with a Gaussian with parameter of $\sigma_I$ (integration scale).

It should be noted that instead of using the DoG-based response, we can try to use some derivatives-based alternatives \cite{mikolajczyk2004scale}. For example, we can define the following gradient-based metric
\begin{equation}\label{grad_metric}
S_{\sigma}(x,y) = g(\sigma_I) \ast \left[\left(\frac{\partial}{\partial x}L(x,y,\sigma)\right)^2+\left(\frac{\partial}{\partial y}L(x,y,\sigma)\right)^2\right],
\end{equation}
or the following scale-normalized metric based on Laplacian of Gaussian (LoG)
\begin{equation}\label{lap_metric}
S_{\sigma}(x,y) = g(\sigma_I) \ast \sigma^2 \left|\frac{\partial^2}{\partial x^2}L(x,y,\sigma) + \frac{\partial^2}{\partial y^2}L(x,y,\sigma) \right|.
\end{equation}
Here we use the DoG-based metric (\ref{dog_metric}) in that the DoG operator is a close approximation of the LoG function but can significantly accelerate the computation process \cite{lowe2004distinctive}. Another kind of possible alternatives is based on the eigenvalues of the second moment matrix \cite{mikolajczyk2004scale,zhou2014multi} or Hessian matrix \cite{lowe2004distinctive}, but it is far more complicated and less stable \cite{mikolajczyk2002detection}.

\begin{figure}[!ht]
	\centering
 	\includegraphics[width=0.8\linewidth]{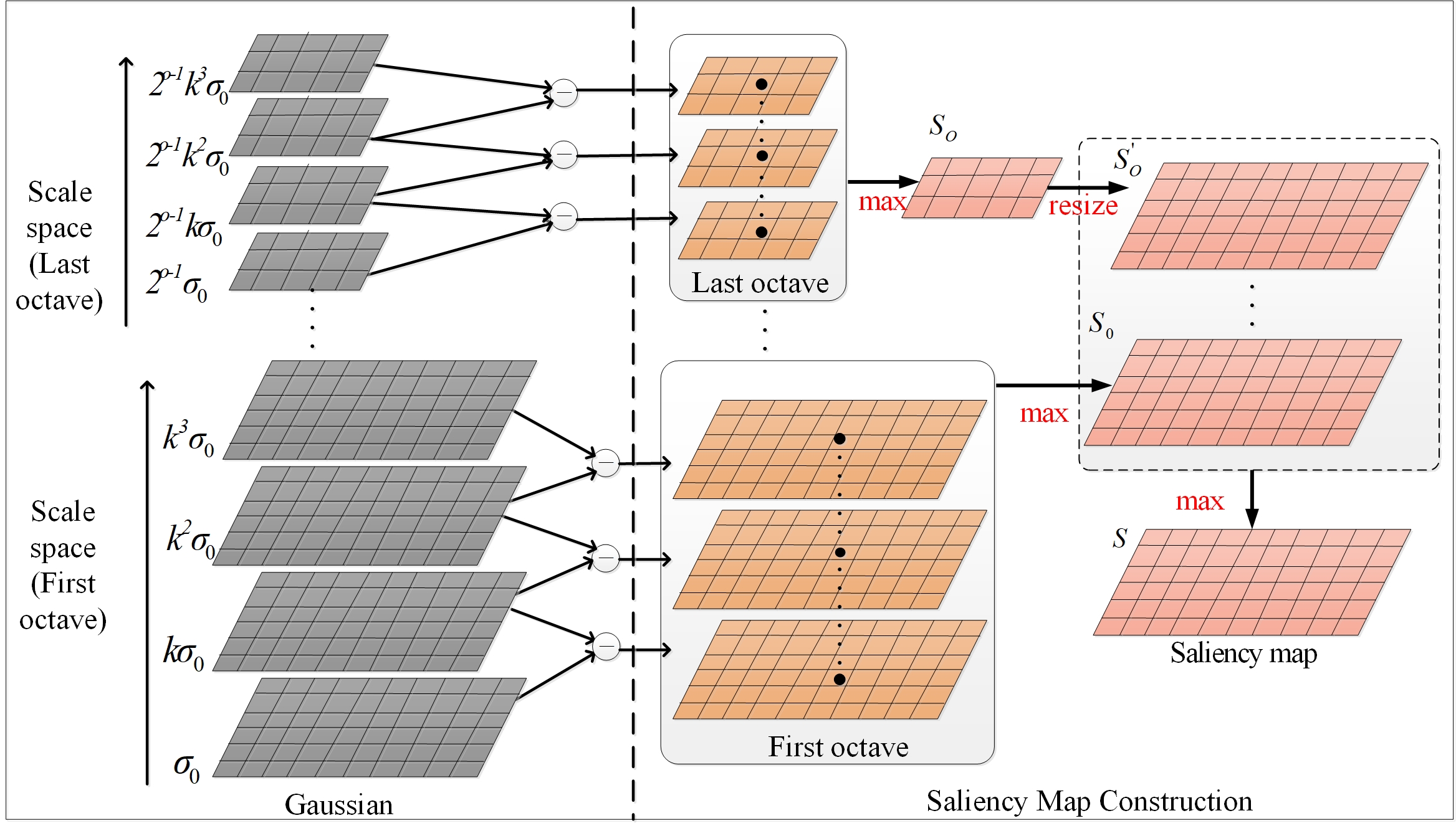}
	\caption{Construction of the scale-invariant saliency map for each image.}
	\label{saliency}
\end{figure}

To construct the scale-invariant saliency map $S(x,y)$ of image $I(x,y)$, for each position, we search the maximum saliency metric across the scale space
\begin{equation}\label{sal}
S(x, y) = \max_\sigma S_\sigma(x,y).
\end{equation}
Due to the image resolution of each octave is different, we first apply max operation within each octave and then resize the resulting map to be exactly the size of the original image. The scale-invariant saliency map is finally obtained by applying the max operation across each octave, as shown in Figure \ref{saliency}.

In our implementation, when generating the DoG pyramid, the number of octaves is $o$ and we set initial scale $\sigma_0 = 1$ and then produce $s+1$ images in the stack of Gaussian blurred images for each octave so that the saliency comparison per octave covers a complete octave. The integration filter $g(\sigma_I)$ is a Gaussian low-pass spatial filter of size $3\times3$.

\subsection{Activity Maps Generation and Fusion}
A straightforward approach is to take the obtained scale-invariant saliency of each pixels as the corresponding \emph{activity} or weight. However, it will introduce blur in the fused image. We adopt the simple \emph{non-max suppression} to alleviate this problem, i.e. for $i$-th image $I^{i}(x,y)$, we can determine a mask $M^{i}(x,y)$ as the initial activity map by comparing the obtained scale-invariant saliency maps $\{S^{i}(x,y)\}_{i=1}^n$
\begin{equation}\label{}
  M^{i}(x,y) = \left\{
                \begin{array}{cl}
                   1, & \mathrm{if}~ S^{i}(x,y) = \max \left(S^{1}(x,y), S^{2}(x,y), \cdots, S^{n}(x,y) \right) \\
                   0, & \mathrm{otherwise,} \\
                 \end{array}
                 \right.
\end{equation}
where $n$ is the number of input images. However, as the above procedure compares pixels individually without considering the spatial context information, the resulting masks are usually noisy and will introduce artifacts into the fused image. Moreover, there may exist more than one maximum at a spatial position in scale-space (i.e., there exist multi-scale structures). To deal with these situations, we can choose a sophisticated solution is to model the pixel saliency and spatial smoothness simultaneously into a energy function, which can be globally optimized by some tools such as graph-cut techniques \cite{kolmogorov2004energy}, but the optimization is often relatively inefficient. Another choice is to perform the morphology smoothing operation which is very efficient but inaccurate which is likely to introduce errors or artifacts \cite{zhou2014multi}. Guided image filtering \cite{he2013guided} or joint bilateral filtering \cite{petschnigg2004digital} is an interesting alternative, which provides a trade-off between the efficiency and accuracy. Following \cite{li2013image}, we determine the final activity map $W(x,y)$ by applying guided filtering on the initial activity map as follows
\begin{equation}\label{gif}
  W(x,y) =  \frac{1}{|\omega|} \sum_{(u,v) \in \omega(x,y)}[ a(u,v) I(x,y) + b(u,v)],
\end{equation}
where $|\omega|$ is the number of pixels in window $\omega(x,y)$ with size of $(2r+1)\times(2r+1)$ which is centered at pixel $(x,y)$, $a(x,y)$ and $b(x,y)$ are the const coefficients of window $\omega(x,y)$ which are determined by \emph{ridge regression}
\begin{eqnarray}
   a(x,y) &=& \frac{\frac{1}{|\omega|} \sum_{(u,v) \in \omega(x,y)} I(u,v) M(u,v) - \mu(x,y) \bar{M}(x,y)}{\delta^2(x,y) + \epsilon} \\
   b(x,y) &=&  \bar{M}(x,y) - a(x,y)\mu(x,y).
\end{eqnarray}
Here $\mu(x,y)$ and $\delta^2(x,y)$ are the mean and variance of image $I$ in window $\omega(x,y)$, $\bar{M}(x,y)$ is the mean of the initial activity map $M$ in window $\omega(x,y)$ and $\epsilon$ denotes the regularization parameter penalizing large $a(x,y)$. The parameters of guided filtering are set to $r=2, \epsilon = 2$ in our implementation.

For each input image $I^i(x,y)$, we can determine the corresponding activity map $W^i(x,y)$ and then obtain the final fused image $F(x,y)$ by
\begin{equation}\label{fuse}
  F(x,y) = \frac{\sum_{i=0}^{n}W^i(x,y) I^i(x,y)}{\sum_{i=0}^{n}W^i(x,y)}.
\end{equation}
For color input images, the activity map is repeated for red, green and blue channels respectively to generate the final color fused image.

\begin{figure}
  \centering
      \begin{subfigure}[b]{0.6\textwidth}
        \includegraphics[width=0.223\textwidth]{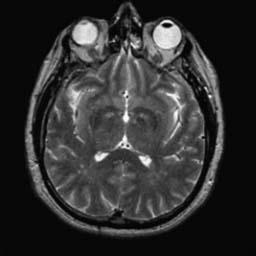}
        \includegraphics[width=0.223\textwidth]{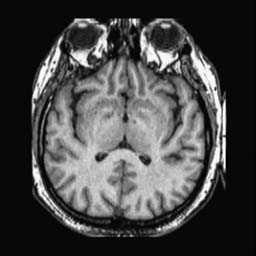}
        \includegraphics[width=0.223\textwidth]{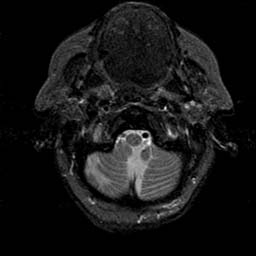}
        \includegraphics[width=0.223\textwidth]{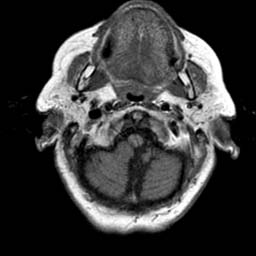}
     \caption{Multi-modal medical images}  \label{fig:multi_model}
    \end{subfigure}
    \begin{subfigure}[b]{0.6\linewidth}
        \includegraphics[width=0.223\linewidth]{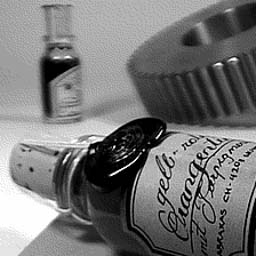}
        \includegraphics[width=0.223\linewidth]{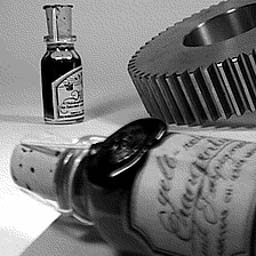}
        \includegraphics[width=0.223\linewidth,height=0.223\linewidth]{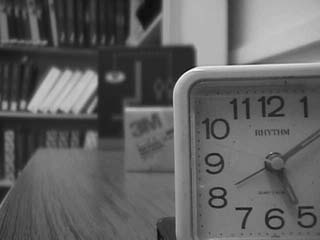}
        \includegraphics[width=0.223\linewidth,height=0.223\linewidth]{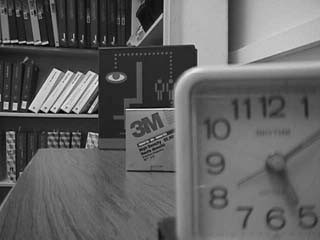}
         \caption{Natural multi-focus images }  \label{fig:natual}
    \end{subfigure}
    \begin{subfigure}[b]{0.6\linewidth}
        \includegraphics[width=0.3\linewidth]{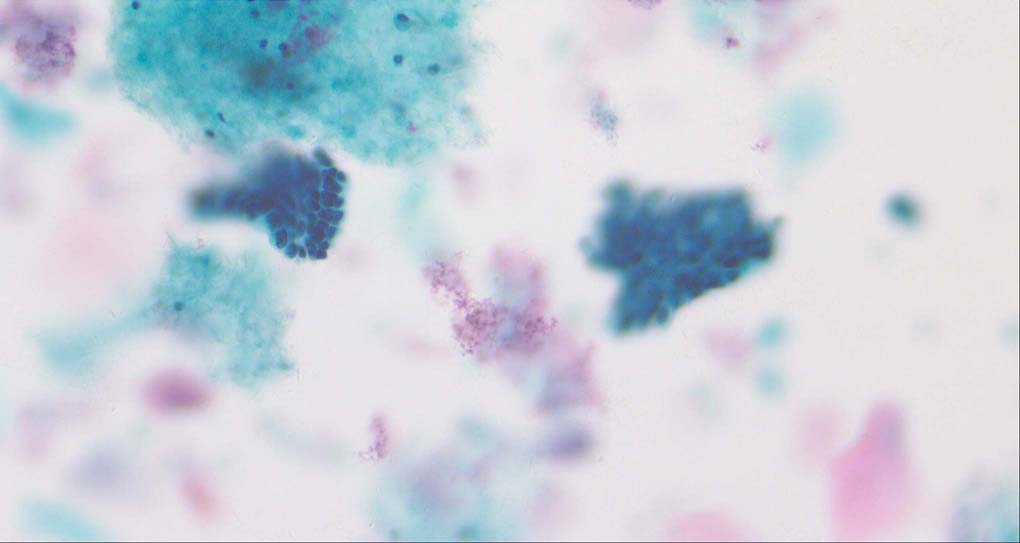}
        \includegraphics[width=0.3\linewidth]{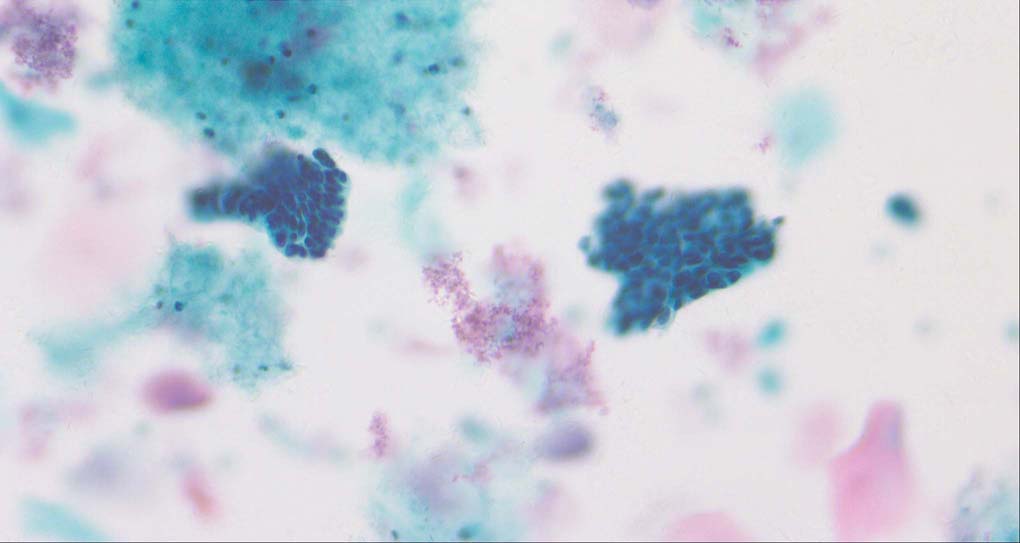}
        \includegraphics[width=0.3\linewidth]{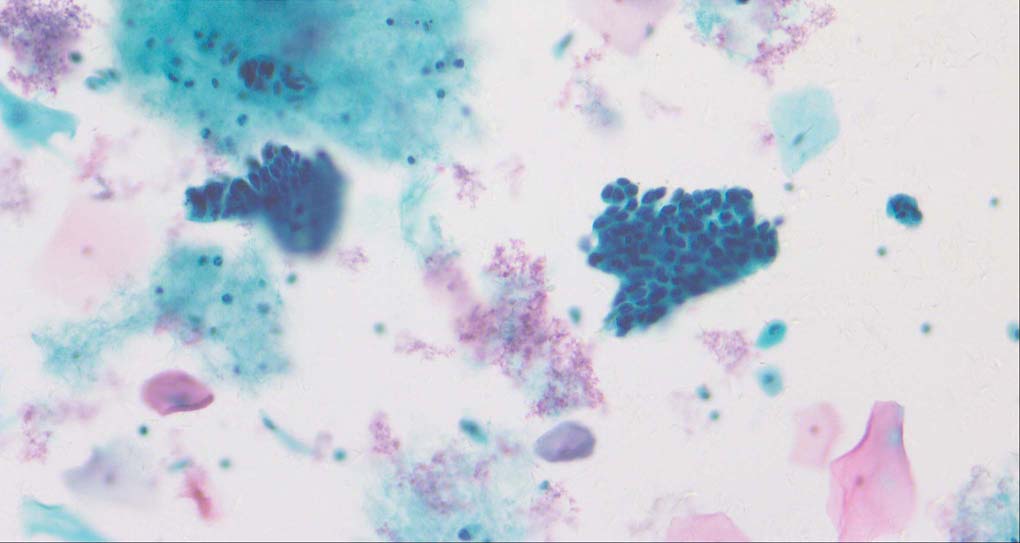}
        \\
        \includegraphics[width=0.3\linewidth]{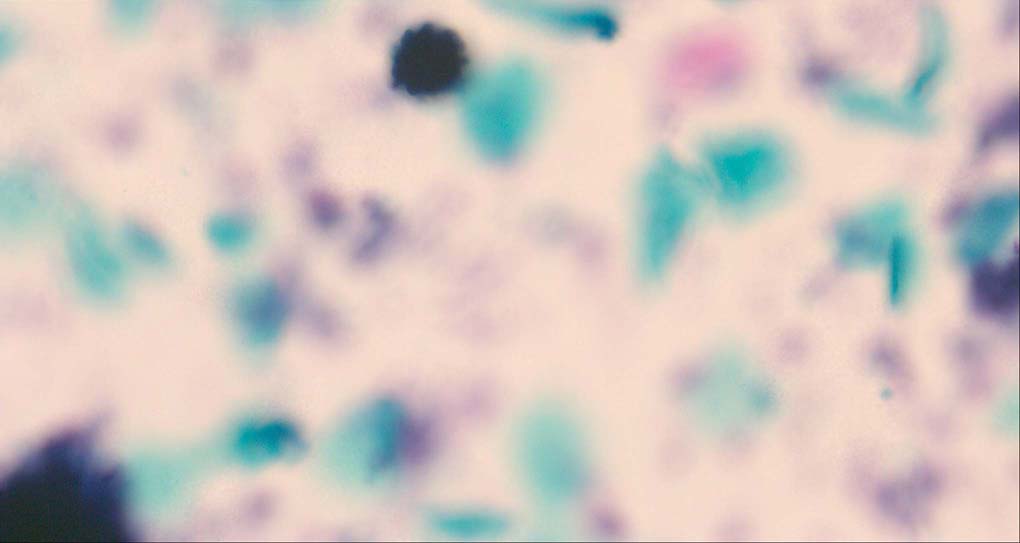}
        \includegraphics[width=0.3\linewidth]{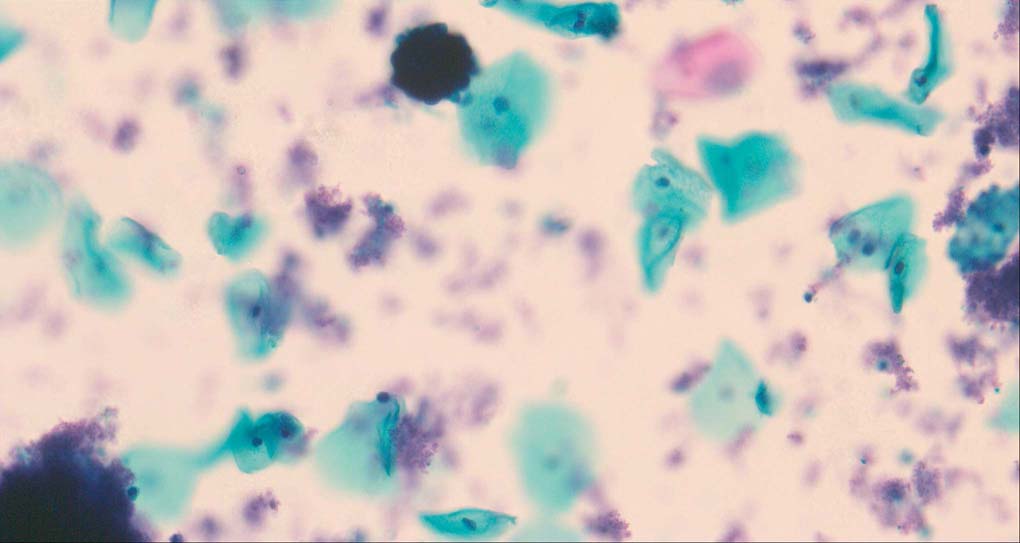}
        \includegraphics[width=0.3\linewidth]{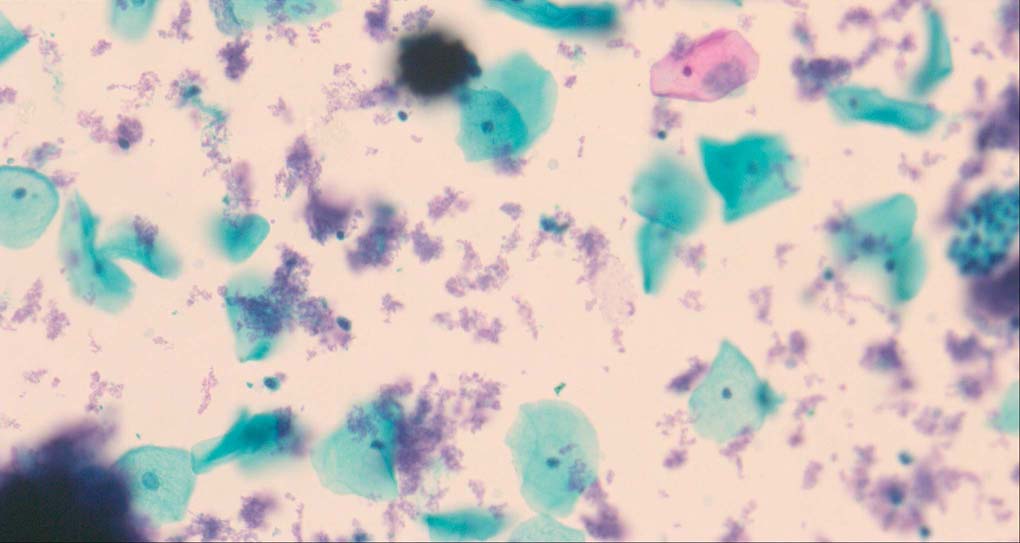}
        \\
        \includegraphics[width=0.3\linewidth]{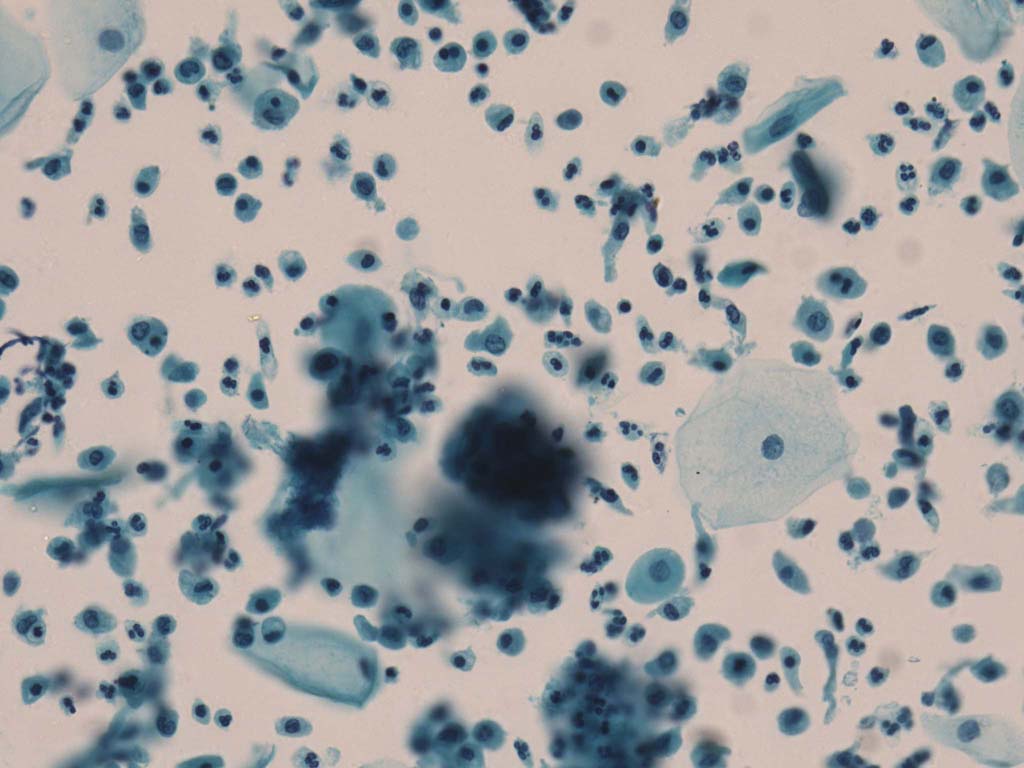}
        \includegraphics[width=0.3\linewidth]{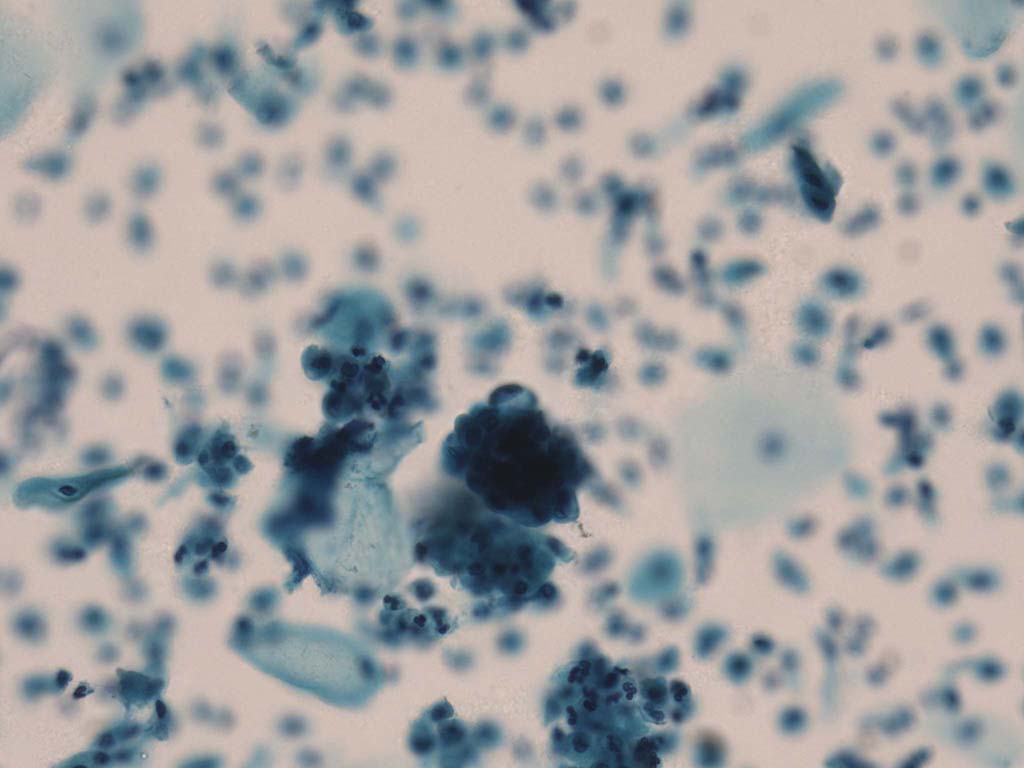}
        \includegraphics[width=0.3\linewidth]{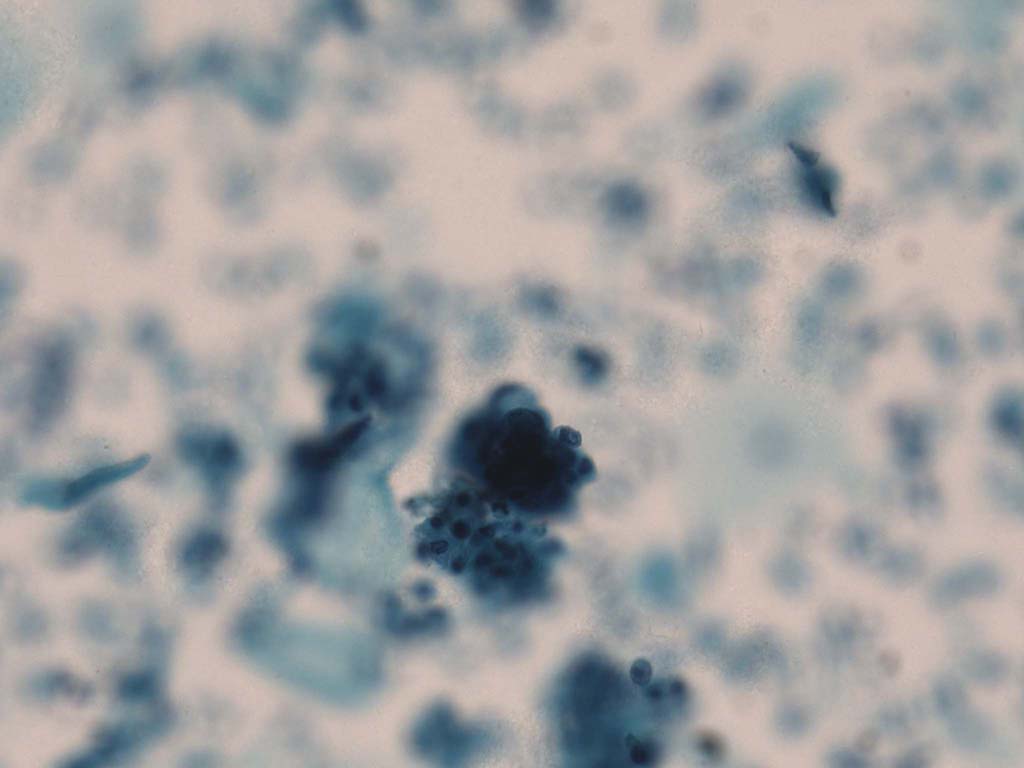}
    \caption{Multi-focus cell images }  \label{fig:cell}
    \end{subfigure}
     \caption{Example source images.}\label{pubdata}
\end{figure}

\section{ Experiments and Discussions}
To demonstrate the effectiveness and efficiency of the proposed image fusion method , we conduct a set of comparative experiments on three image datasets. The first is composed by 8 pairs of multi-modal medical images and the second one contains 15 pairs of multi-focus gray or color natural images. These two datasets are often used in many related papers and some examples are shown in Figure \ref{fig:multi_model} and Figure \ref{fig:natual}. The third one is a new multi-focus cervical cell image dataset collected by ourselves, which consists of 15 groups of color images and each group contains a series of multi-focus cervix cell images with size of $2040 \times 1086$ or $2448 \times 2048$, etc. Some source examples are shown in Figure \ref{fig:cell}. Our source code implemented in C++ along with the new multi-focus cervical cell image dataset is available online.

We compare our method with several state-of-the-art algorithms, which include dense SIFT (DSIFT)\cite{liu2015multi}, dual-tree complex wavelet transform (DTCWT) \cite{zhou2014multi}, guided filter fusion (GFF) \cite{li2013image}, image matting (IM) \cite{li2013image2}, CNN-based method \cite{Liu2017Multi}, Laplacian sparse representation (LP-SR) \cite{liu2015general}, multi-weighted gradient fusion (MWGF) \cite{zhou2014multi}, non-sampled contourlet transform (NSCT) \cite{zhang2009multifocus} and boundary finding based method (BF) \cite{Zhang2017Boundary}. All the methods are implemented in Matlab except the CNN-based method which is implemented with C++ based on caffe \cite{jia2014caffe} and the parameters of them are set with default values given by the authors. The results are compared in terms of both visual and objective quality. For the objective quality evaluation of image fusion, we adopt several commonly metrics, including the mutual information (MI) \cite{Qu2002Information}, structural similarity (SSIM) \cite{wang2004image}, quality index (QI) \cite{wang2002universal}, edge information preservation value ($\mathrm{Q}_{\mathrm{AB/F}}$) \cite{xydeas2000objective}, feature mutual information (FMI) \cite{haghighat2011non} and visual information fidelity (VIF) \cite{sheikh2006image}.
\begin{figure}
    \centering
    \begin{minipage}[t]{\textwidth}
    \hspace{0.12\textwidth}$s=1$    \hspace{0.055\textwidth}$s=2$    \hspace{0.055\textwidth}$s=3$ \hspace{0.055\textwidth}$s=4$    \hspace{0.055\textwidth}$s=5$    \hspace{0.055\textwidth}$s=6$
    \hspace{0.055\textwidth}$s=7$    \hspace{0.055\textwidth}$s=8$
    \end{minipage}
    \rotatebox{90}{\hspace{0.02\textwidth}$o=1$}
        \includegraphics[width=0.1\textwidth]{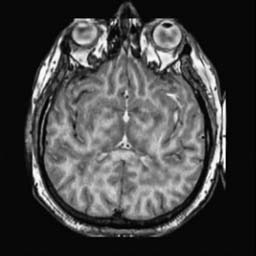}
        \includegraphics[width=0.1\textwidth]{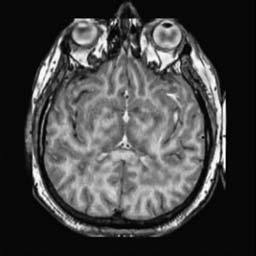}
        \includegraphics[width=0.1\textwidth]{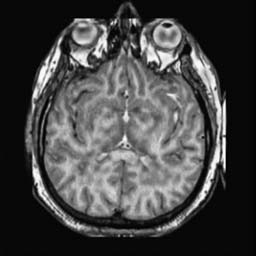}
        \includegraphics[width=0.1\textwidth]{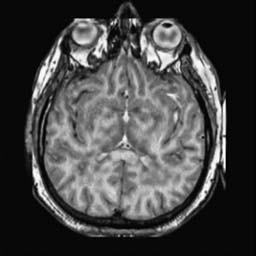}
        \includegraphics[width=0.1\textwidth]{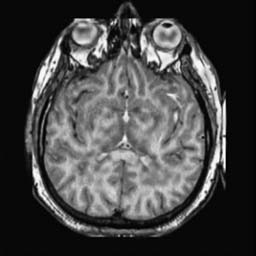}
        \includegraphics[width=0.1\textwidth]{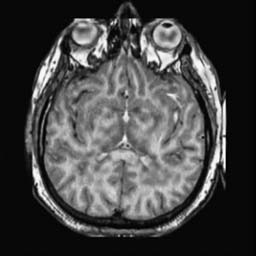}
        \includegraphics[width=0.1\textwidth]{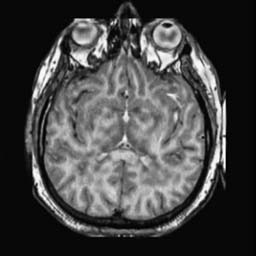}
        \includegraphics[width=0.1\textwidth]{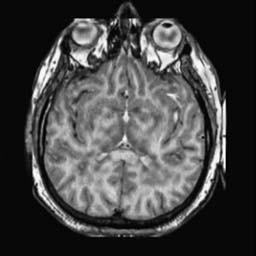}\\
      \rotatebox{90}{\hspace{0.02\textwidth}$o=2$}
        \includegraphics[width=0.1\textwidth]{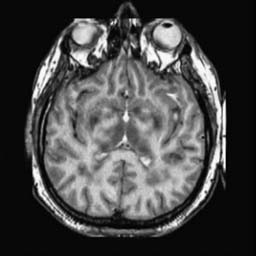}
        \includegraphics[width=0.1\textwidth]{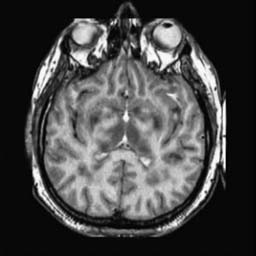}
        \includegraphics[width=0.1\textwidth]{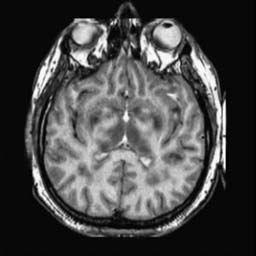}
        \includegraphics[width=0.1\textwidth]{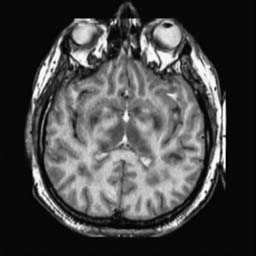}
        \includegraphics[width=0.1\textwidth]{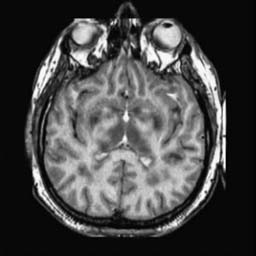}
        \includegraphics[width=0.1\textwidth]{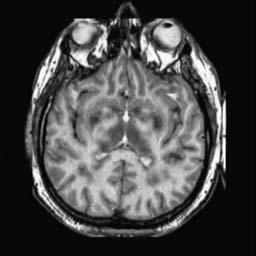}
        \includegraphics[width=0.1\textwidth]{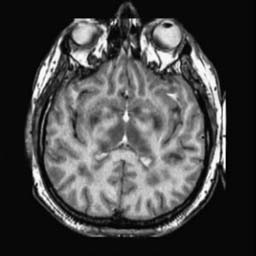}
        \includegraphics[width=0.1\textwidth]{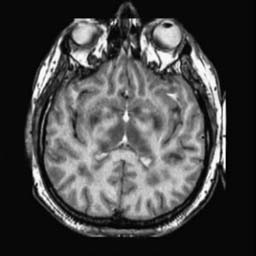}\\
    \rotatebox{90}{\hspace{0.02\textwidth}$o=3$}
        \includegraphics[width=0.1\textwidth]{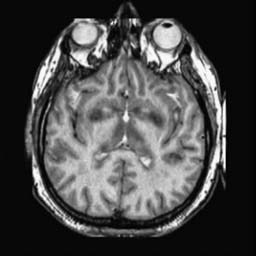}
        \includegraphics[width=0.1\textwidth]{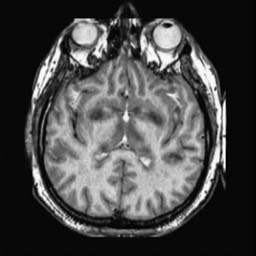}
        \includegraphics[width=0.1\textwidth]{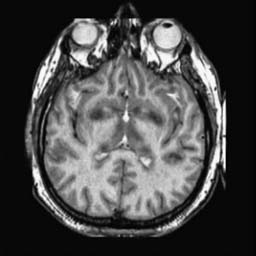}
        \includegraphics[width=0.1\textwidth]{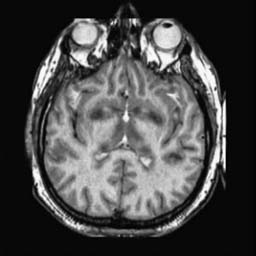}
        \includegraphics[width=0.1\textwidth]{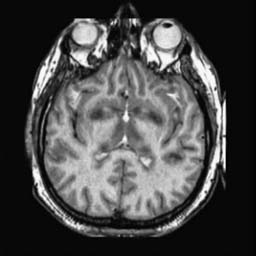}
        \includegraphics[width=0.1\textwidth]{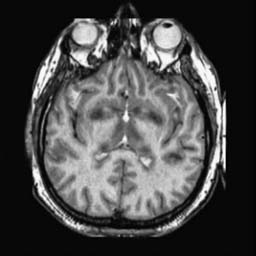}
        \includegraphics[width=0.1\textwidth]{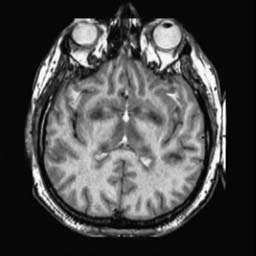}
        \includegraphics[width=0.1\textwidth]{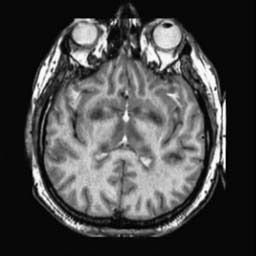}\\
    \rotatebox{90}{\hspace{0.02\textwidth}$o=4$}
        \includegraphics[width=0.1\textwidth]{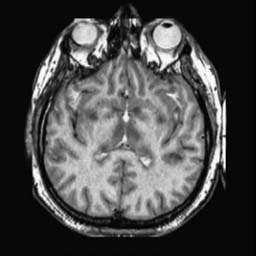}
        \includegraphics[width=0.1\textwidth]{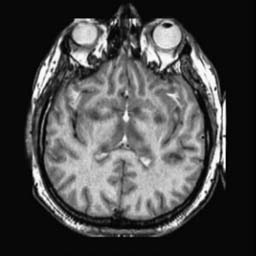}
        \includegraphics[width=0.1\textwidth]{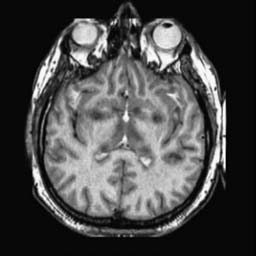}
        \includegraphics[width=0.1\textwidth]{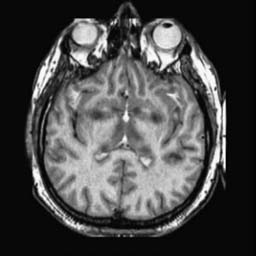}
        \includegraphics[width=0.1\textwidth]{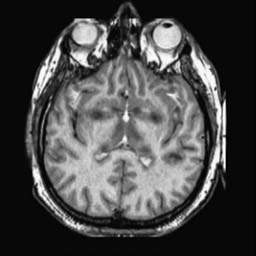}
        \includegraphics[width=0.1\textwidth]{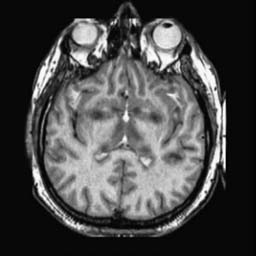}
        \includegraphics[width=0.1\textwidth]{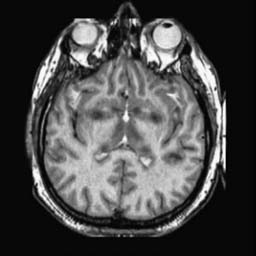}
        \includegraphics[width=0.1\textwidth]{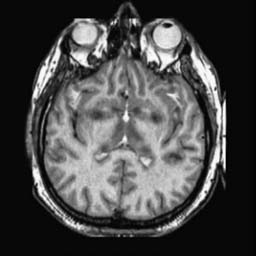}\\
    \rotatebox{90}{\hspace{0.02\textwidth}$o=5$}
        \includegraphics[width=0.1\textwidth]{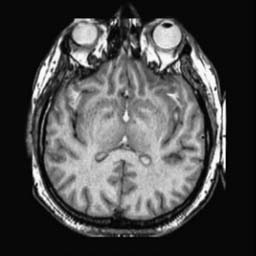}
        \includegraphics[width=0.1\textwidth]{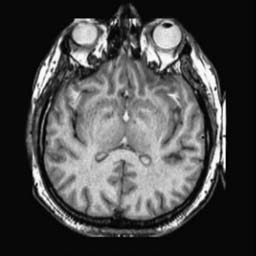}
        \includegraphics[width=0.1\textwidth]{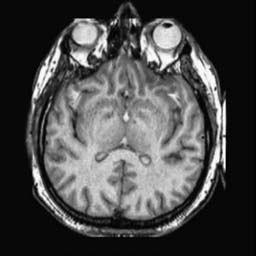}
        \includegraphics[width=0.1\textwidth]{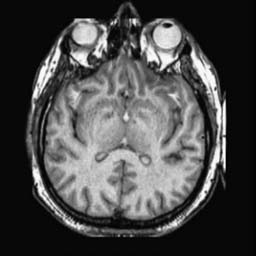}
        \includegraphics[width=0.1\textwidth]{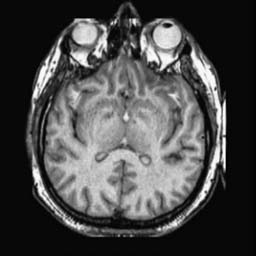}
        \includegraphics[width=0.1\textwidth]{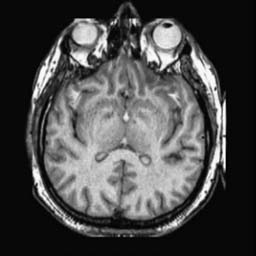}
        \includegraphics[width=0.1\textwidth]{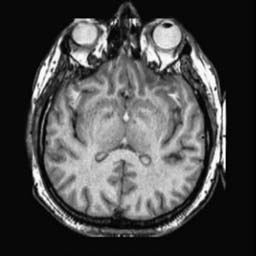}
        \includegraphics[width=0.1\textwidth]{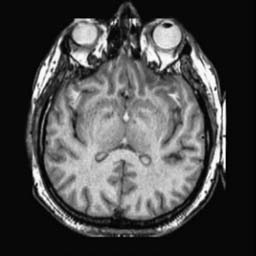}\\
    \caption{The fused image obtained by the proposed method with varying $o$ and $s$. Notice that the integrity information of large size objects (e.g. eyeballs) is reserved  by increasing the value $o$ and $s$.}\label{fused_result_with_varing_params}
\end{figure}
\begin{figure}
\centering
    \begin{subfigure}[b]{0.48\textwidth}
        \includegraphics[width=\textwidth]{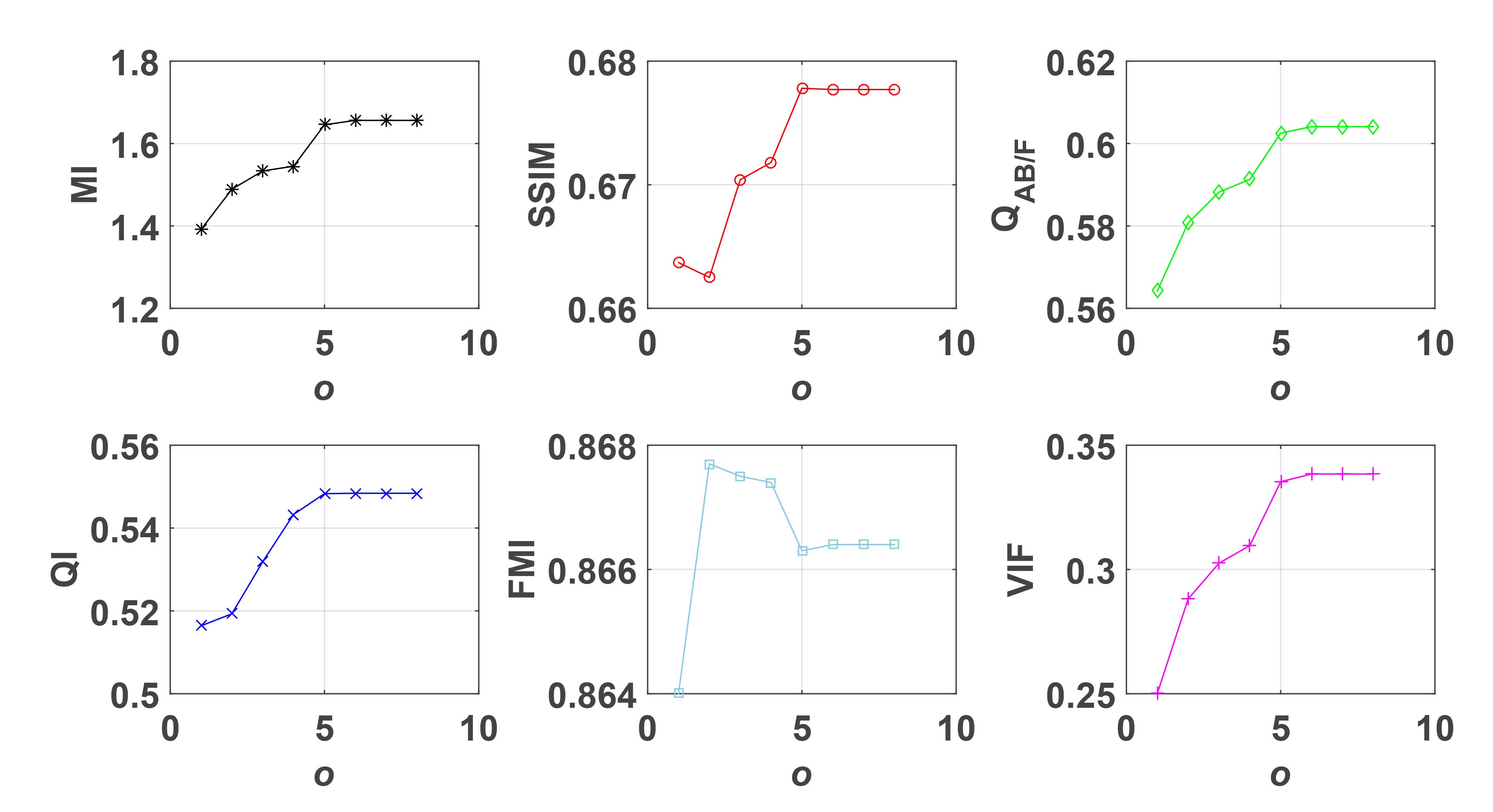}\caption{$s=3$}\label{o:change}
    \end{subfigure}
    \begin{subfigure}[b]{0.48\textwidth}
        \includegraphics[width=\textwidth]{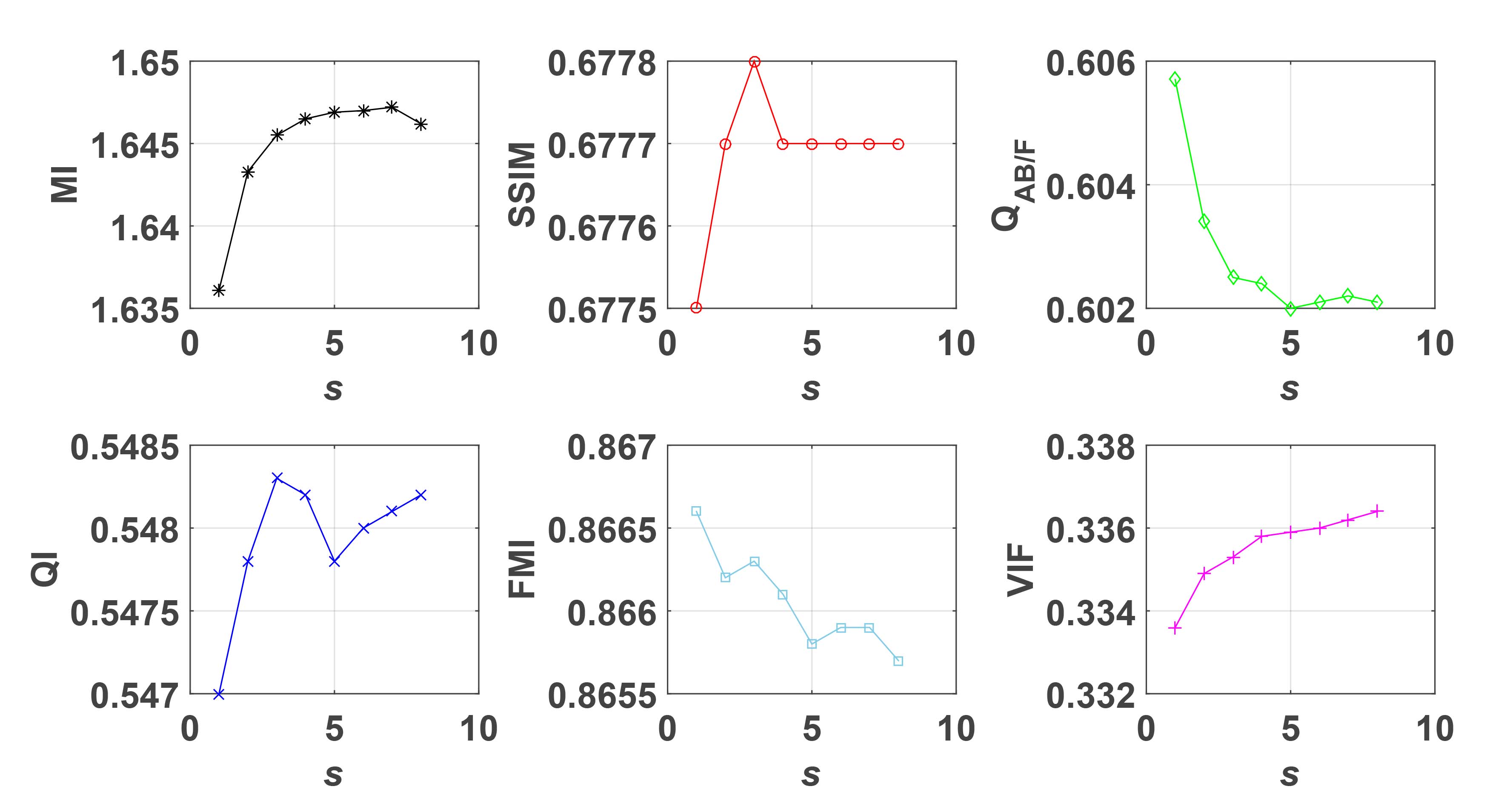}\caption{$o=5$}\label{s:change}
    \end{subfigure}
    \caption{Performance of the proposed method with varying $o$ and $s$.}\label{performance_varying_params}
\end{figure}
\begin{figure}
    \begin{minipage}[t]{0.5\textwidth}
    \begin{subfigure}[b]{0.2\textwidth}
        \includegraphics[width=\textwidth]{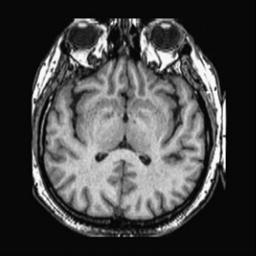}
        \caption{DSIFT}       \label{fig12:DSIFT}
    \end{subfigure}
    \begin{subfigure}[b]{0.2\textwidth}
        \includegraphics[width=\textwidth]{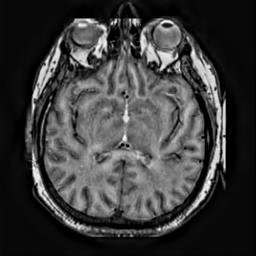}
        \caption{DTCWT}        \label{fig12:DTCWT}
    \end{subfigure}
    \begin{subfigure}[b]{0.2\textwidth}
        \includegraphics[width=\textwidth]{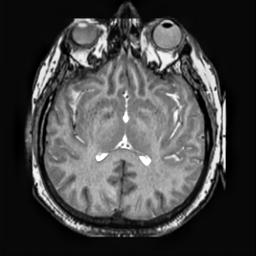}
        \caption{GFF}         \label{fig12:gff}
    \end{subfigure}
    \begin{subfigure}[b]{0.2\textwidth}
        \includegraphics[width=\textwidth]{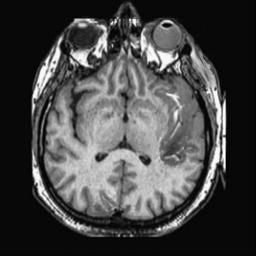}
        \caption{IM}        \label{fig12:im}
    \end{subfigure}\\
    \begin{subfigure}[b]{0.2\textwidth}
        \includegraphics[width=\textwidth]{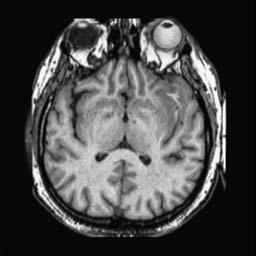}
        \caption{CNN}        \label{fig12:CNN}
    \end{subfigure}
    \begin{subfigure}[b]{0.2\textwidth}
        \includegraphics[width=\textwidth]{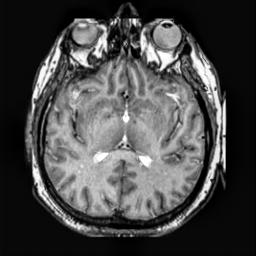}
        \caption{LP-SR}        \label{fig12:LPSR}
    \end{subfigure}
    \begin{subfigure}[b]{0.2\textwidth}
        \includegraphics[width=\textwidth]{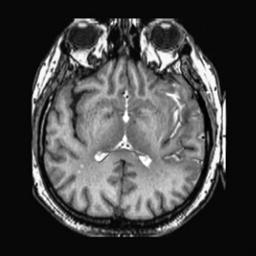}
        \caption{MWGF}        \label{fig12:MWGF}
    \end{subfigure}
    \begin{subfigure}[b]{0.2\textwidth}
        \includegraphics[width=\textwidth]{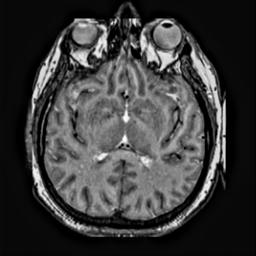}
        \caption{NSCT}        \label{fig12:NSCT}
    \end{subfigure}\\
    \begin{subfigure}[b]{0.2\textwidth}
        \includegraphics[width=\textwidth]{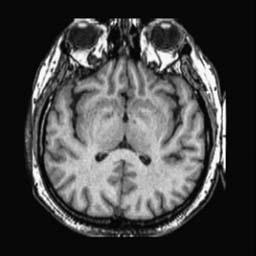}
        \caption{BF}        \label{fig12:BF}
    \end{subfigure}
    \begin{subfigure}[b]{0.2\textwidth}
        \includegraphics[width=\textwidth]{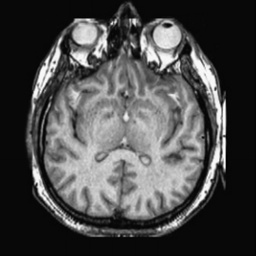}
        \caption{Proposed}       \label{fig12:ours}
    \end{subfigure}
   \end{minipage}
   \begin{minipage}[t]{0.5\textwidth}
    \begin{subfigure}[b]{0.2\textwidth}
        \includegraphics[width=\textwidth]{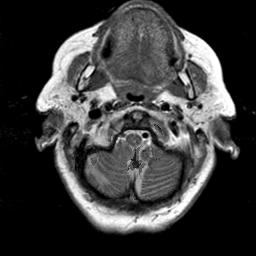}
        \caption{DSIFT}
        \label{fig13:DSIFT}
    \end{subfigure}
    \begin{subfigure}[b]{0.2\textwidth}
        \includegraphics[width=\textwidth]{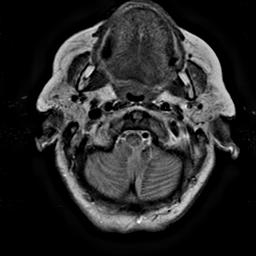}
        \caption{DTCWT}
        \label{fig13:DTCWT}
    \end{subfigure}
    \begin{subfigure}[b]{0.2\textwidth}
        \includegraphics[width=\textwidth]{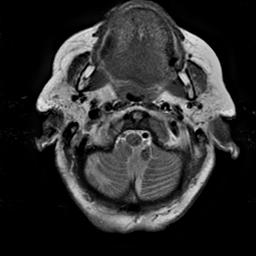}
        \caption{GFF}
        \label{fig13:gff}
    \end{subfigure}
    \begin{subfigure}[b]{0.2\textwidth}
        \includegraphics[width=\textwidth]{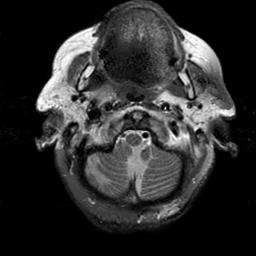}
        \caption{IM}
        \label{fig13:im}
    \end{subfigure}\\
    \begin{subfigure}[b]{0.2\textwidth}
        \includegraphics[width=\textwidth]{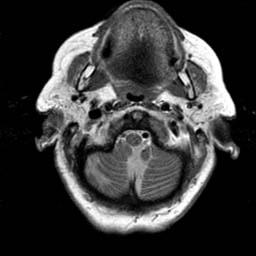}
        \caption{CNN}
        \label{fig13:CNN}
    \end{subfigure}
    \begin{subfigure}[b]{0.2\textwidth}
        \includegraphics[width=\textwidth]{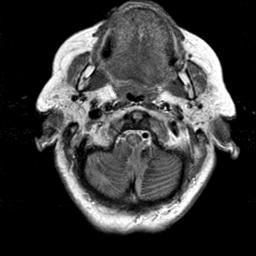}
        \caption{LP-SR}
        \label{fig13:LPSR}
    \end{subfigure}
    \begin{subfigure}[b]{0.2\textwidth}
        \includegraphics[width=\textwidth]{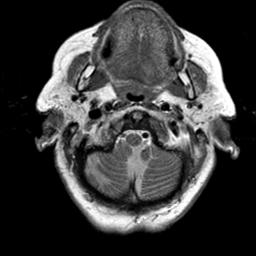}
        \caption{MWGF}
        \label{fig13:MWGF}
    \end{subfigure}
    \begin{subfigure}[b]{0.2\textwidth}
        \includegraphics[width=\textwidth]{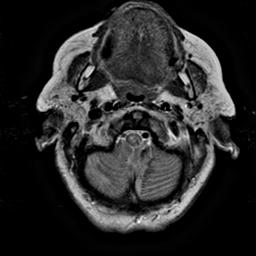}
        \caption{NSCT}
        \label{fig13:NSCT}
    \end{subfigure}\\
    \begin{subfigure}[b]{0.2\textwidth}
        \includegraphics[width=\textwidth]{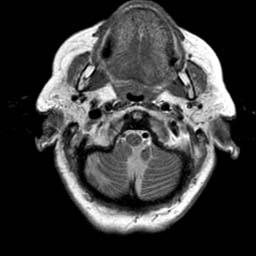}
        \caption{BF}
        \label{fig13:BF}
    \end{subfigure}
    \begin{subfigure}[b]{0.2\textwidth}
        \includegraphics[width=\textwidth]{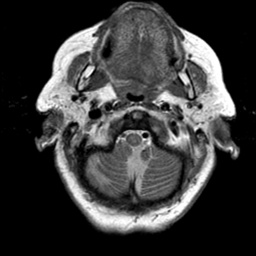}
        \caption{Proposed}
        \label{fig13:ours}
    \end{subfigure}
   \end{minipage}
  \caption{The fused results obtained by different methods for the multi-modal medical images.}\label{multimodel1_result}
\end{figure}

We first evaluate the performance of the proposed method under varying total number of octaves $o$ and number of layers $s$ sampled per octave. The fused images of a pair of multi-modal medical images with different $o$ and $s$ are shown in Figure \ref{fused_result_with_varing_params}. In this example, on the one hand, when only 1 or 2 octaves are involved in constructing the DoG pyramid, the fused images fail to keep the integrity information of large size objects (e.g. eyeballs), while by increasing the value of $o$, the integrity information of eyeballs is preserved. On the other hand, although not as significant as the increase of octave numbers $o$, the fused image can contain more details by the increase of layer numbers $s$. The corresponding objective quality metrics are shown in Figure \ref{performance_varying_params}. As shown in Figure \ref{o:change}, most of the metric values are improved as the number of octaves increases with the fixed layer numbers 3 in the global tendency and each of them tends to be stable when the number of octaves is 5. To get a relatively good quality from Figure \ref{s:change}, we can notice that some of the metric values can get a good performance when the number of layers is 3, such as the MI, SSIM, QI and VIF, though there are only a little change of all the metric values by increasing the number of layers with the fixed octave numbers 5. Because it will result in more computation burden with the increase of the value $o$ and $s$, and for different kinds of source images, there are different performance with the diverse parameter settings. To get a trade-off between them in our experiments, we set $o=5, s=3$ for the multi-modal dataset, $o=1, s=1$ for the natural datasets and $o=3, s=5$ for the multi-focus cell dataset, respectively.

Figure \ref{multimodel1_result} shows the fused images obtained by different methods with the multi-modal source images shown in Figure \ref{fig:multi_model}. As shown in these figures, the proposed method can produce images which preserve the complementary information of different source images well. Moreover, due to the scale-invariant structure saliency selection, our method can keep the integrity information of large size objects and the visual details simultaneously. Although the fused image generated by other methods can also capture the details to some extent, all of them fail to keep the integrity information of large size objects such as the eyeballs. Furthermore, from Figure \ref{fig13:DSIFT}-\ref{fig13:ours}, the DTCWT, GFF, IM and NSCT methods may decrease the brightness and contrast while the proposed method can preserve these features and details without producing visible artifacts and brightness distortions.

\begin{figure}
    \centering
    \begin{subfigure}[b]{0.18\textwidth}
        \includegraphics[width=\textwidth]{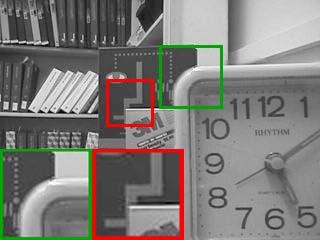}
        \caption{DSIFT}
        \label{fig10:DSIFT}
    \end{subfigure}
    \begin{subfigure}[b]{0.18\textwidth}
        \includegraphics[width=\textwidth]{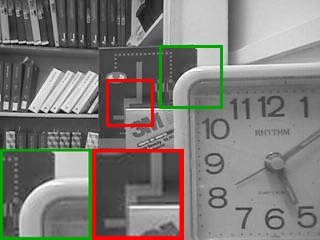}
        \caption{DTCWT}
        \label{fig10:DTCWT}
    \end{subfigure}
    \begin{subfigure}[b]{0.18\textwidth}
        \includegraphics[width=\textwidth]{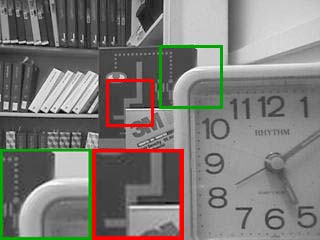}
        \caption{GFF}
        \label{fig10:gff}
    \end{subfigure}
    \begin{subfigure}[b]{0.18\textwidth}
        \includegraphics[width=\textwidth]{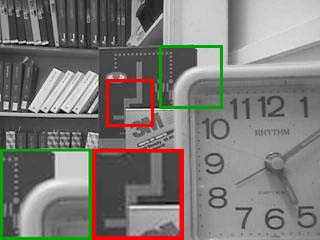}
        \caption{IM}
        \label{fig10:im}
    \end{subfigure}
    \begin{subfigure}[b]{0.18\textwidth}
        \includegraphics[width=\textwidth]{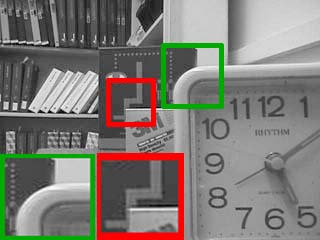}
        \caption{CNN}
        \label{fig10:CNN}
    \end{subfigure} \\
    \begin{subfigure}[b]{0.18\textwidth}
        \includegraphics[width=\textwidth]{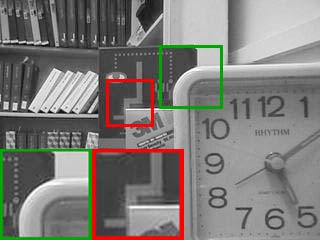}
        \caption{LP-SR}
        \label{fig10:LPSR}
    \end{subfigure}
    \begin{subfigure}[b]{0.18\textwidth}
        \includegraphics[width=\textwidth]{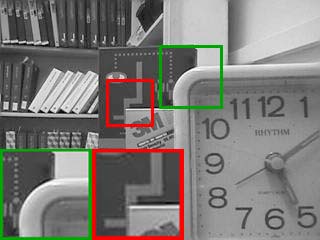}
        \caption{MWGF}
        \label{fig10:MWGF}
    \end{subfigure}
    \begin{subfigure}[b]{0.18\textwidth}
        \includegraphics[width=\textwidth]{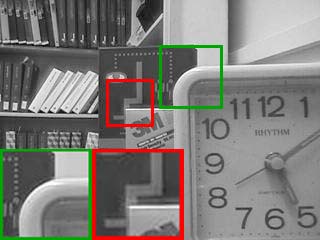}
        \caption{NSCT}
        \label{fig10:NSCT}
    \end{subfigure}
    \begin{subfigure}[b]{0.18\textwidth}
        \includegraphics[width=\textwidth]{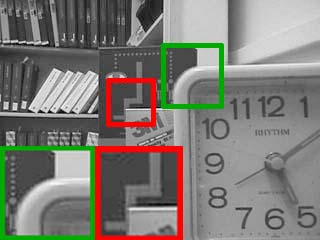}
        \caption{BF}
        \label{fig10:BF}
    \end{subfigure}
    \begin{subfigure}[b]{0.18\textwidth}
        \includegraphics[width=\textwidth]{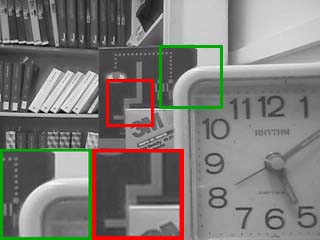}
        \caption{Proposed}
        \label{fig10:ours}
    \end{subfigure}
    \caption{Fusion results of the ``Clock'' source images in Figure \ref{fig:natual}.}\label{watch_results}
\end{figure}

\begin{figure}
    \centering
   \begin{minipage}[t]{0.8\textwidth}
    \begin{subfigure}[b]{0.18\textwidth}
        \includegraphics[width=\textwidth]{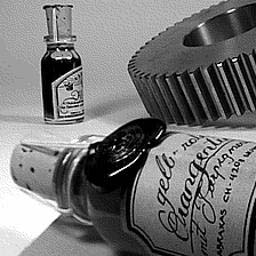}
        \caption{DSIFT}
        \label{Fig10:DSIFT}
    \end{subfigure}
    \begin{subfigure}[b]{0.18\textwidth}
        \includegraphics[width=\textwidth]{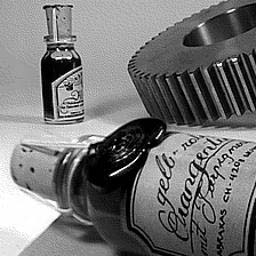}
        \caption{DTCWT}
        \label{Fig10:DTCWT}
    \end{subfigure}
    \begin{subfigure}[b]{0.18\textwidth}
        \includegraphics[width=\textwidth]{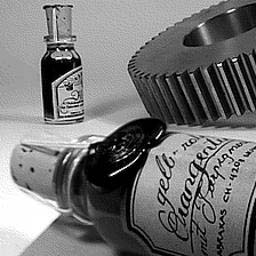}
        \caption{GFF}
        \label{Fig10:gff}
    \end{subfigure}
    \begin{subfigure}[b]{0.18\textwidth}
        \includegraphics[width=\textwidth]{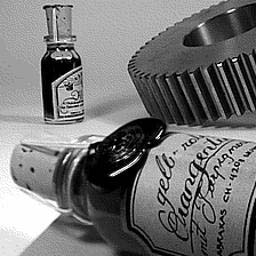}
        \caption{IM}
        \label{Fig10:im}
    \end{subfigure}
    \begin{subfigure}[b]{0.18\textwidth}
        \includegraphics[width=\textwidth]{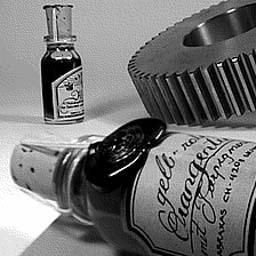}
        \caption{CNN}
        \label{Fig10:CNN}
    \end{subfigure}\\
    \begin{subfigure}[b]{0.18\textwidth}
        \includegraphics[width=\textwidth]{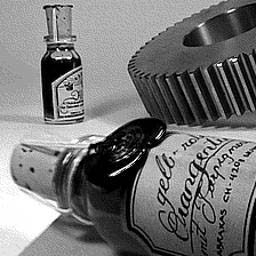}
        \caption{LP-SR}
        \label{Fig10:LPSR}
    \end{subfigure}
    \begin{subfigure}[b]{0.18\textwidth}
        \includegraphics[width=\textwidth]{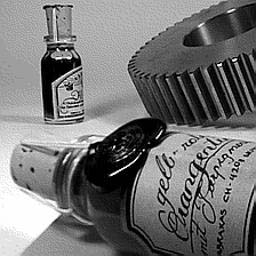}
        \caption{MWGF}
        \label{Fig10:MWGF}
    \end{subfigure}
    \begin{subfigure}[b]{0.18\textwidth}
        \includegraphics[width=\textwidth]{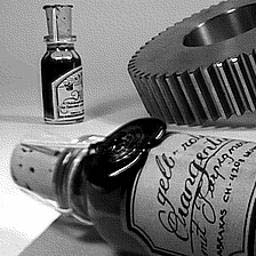}
        \caption{NSCT}
        \label{Fig10:NSCT}
    \end{subfigure}
    \begin{subfigure}[b]{0.18\textwidth}
        \includegraphics[width=\textwidth]{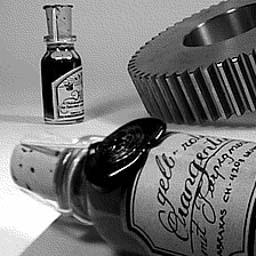}
        \caption{BF}
        \label{Fig10:BF}
    \end{subfigure}
    \begin{subfigure}[b]{0.18\textwidth}
        \includegraphics[width=\textwidth]{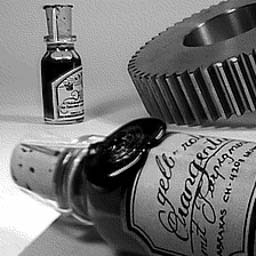}
        \caption{Proposed}
        \label{Fig10:ours}
    \end{subfigure}
    \caption{Fusion results of the ``wine'' source images in Figure \ref{fig:natual}.}\label{wine_results}
   \end{minipage}
\end{figure}

 Figure \ref{watch_results} and Figure \ref{wine_results} show the fused images of different methods by fusing the natural image pairs shown in Figure \ref{fig:natual} respectively. A closeup view is presented in the bottom of each sub-picture in Figure \ref{watch_results}. It can be shown from these figures that although all these methods generate acceptable fused images, our method produces slightly better results than others (see the halo artifacts in the magnified area of Figure \ref{watch_results}).
\begin{figure}
    \centering
    \begin{subfigure}[b]{0.18\textwidth}
        \includegraphics[width=\textwidth]{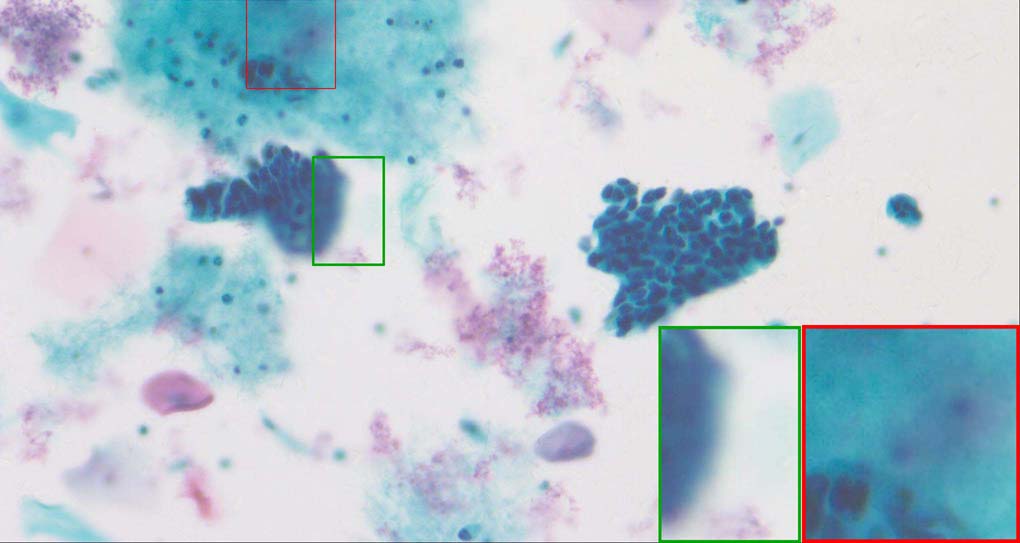}
        \caption{DSIFT}
        \label{figM1:DSIFT}
    \end{subfigure}
    \begin{subfigure}[b]{0.18\textwidth}
        \includegraphics[width=\textwidth]{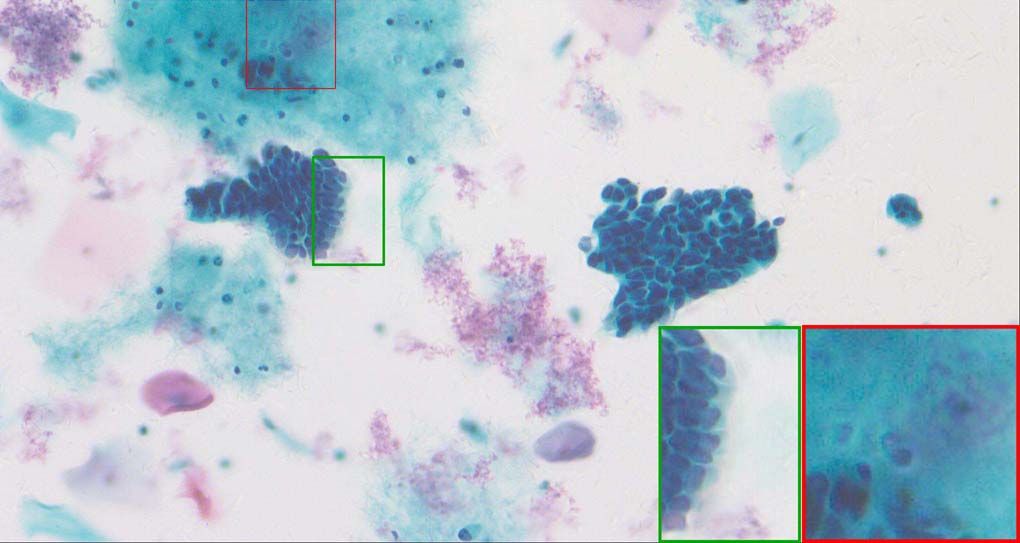}
        \caption{DTCWT}
        \label{figM1:DTCWT}
    \end{subfigure}
    \begin{subfigure}[b]{0.18\textwidth}
        \includegraphics[width=\textwidth]{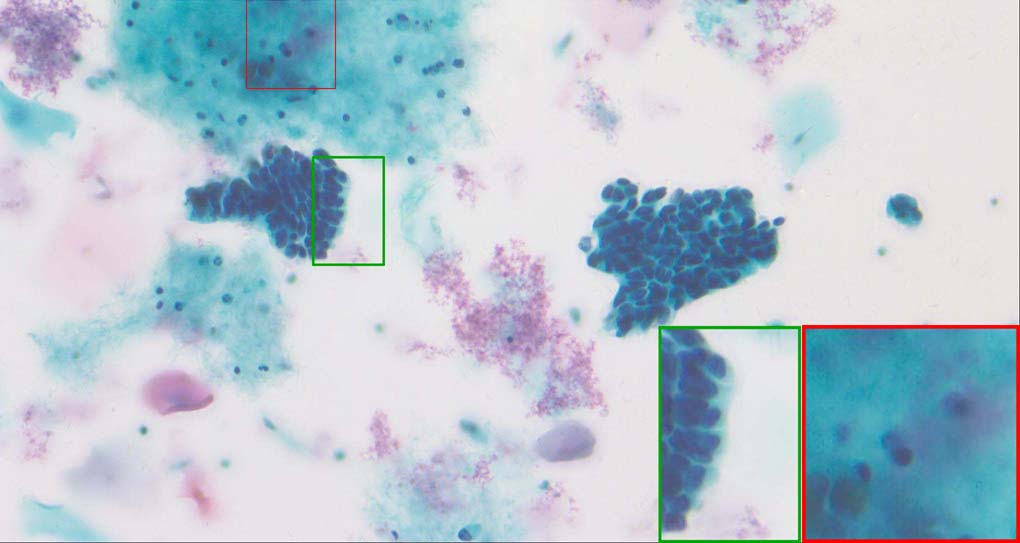}
        \caption{GFF}
        \label{figM1:gff}
    \end{subfigure}
    \begin{subfigure}[b]{0.18\textwidth}
        \includegraphics[width=\textwidth]{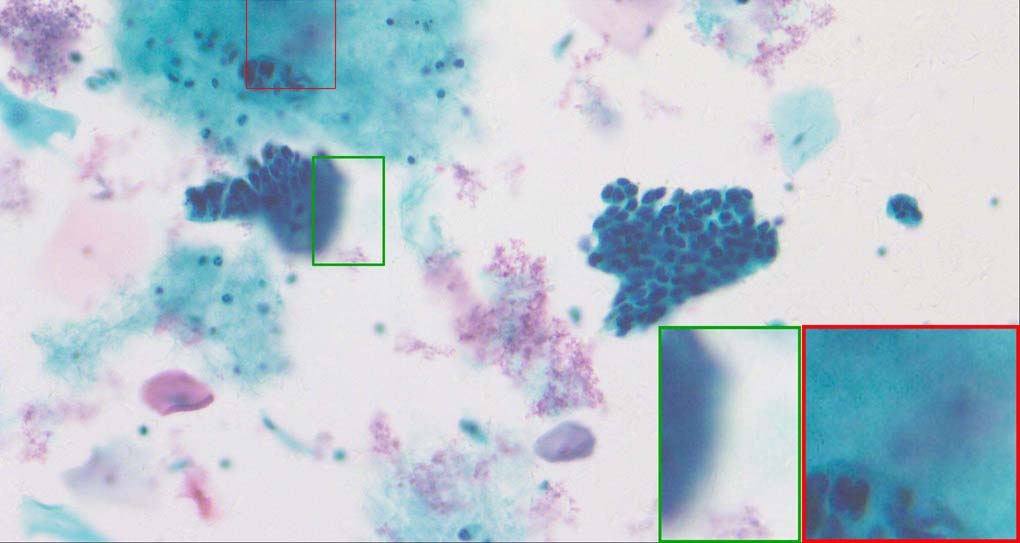}
        \caption{IM}
        \label{figM1:im}
    \end{subfigure}
    \begin{subfigure}[b]{0.18\textwidth}
        \includegraphics[width=\textwidth]{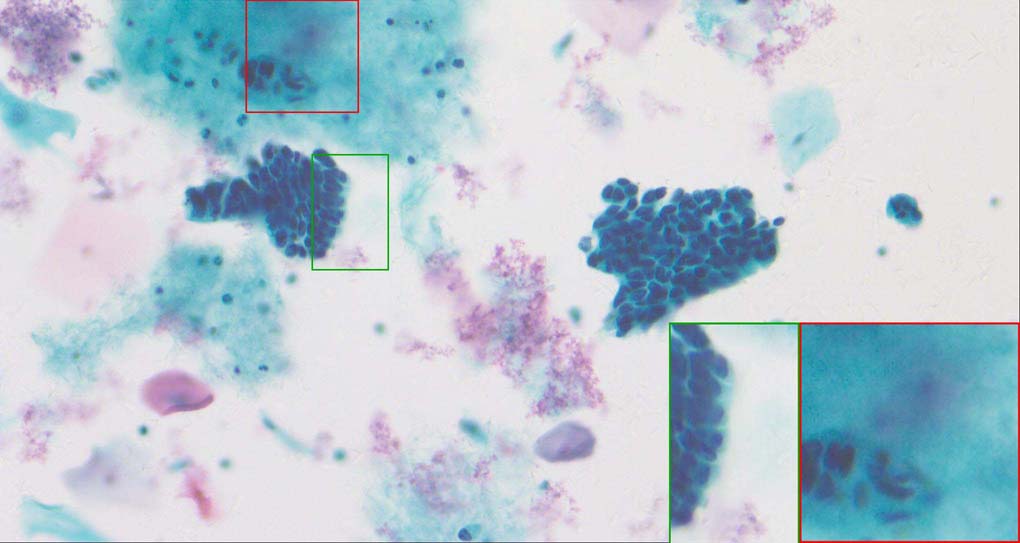}
        \caption{CNN}
        \label{figM1:CNN}
    \end{subfigure}  \\
    \begin{subfigure}[b]{0.18\textwidth}
        \includegraphics[width=\textwidth]{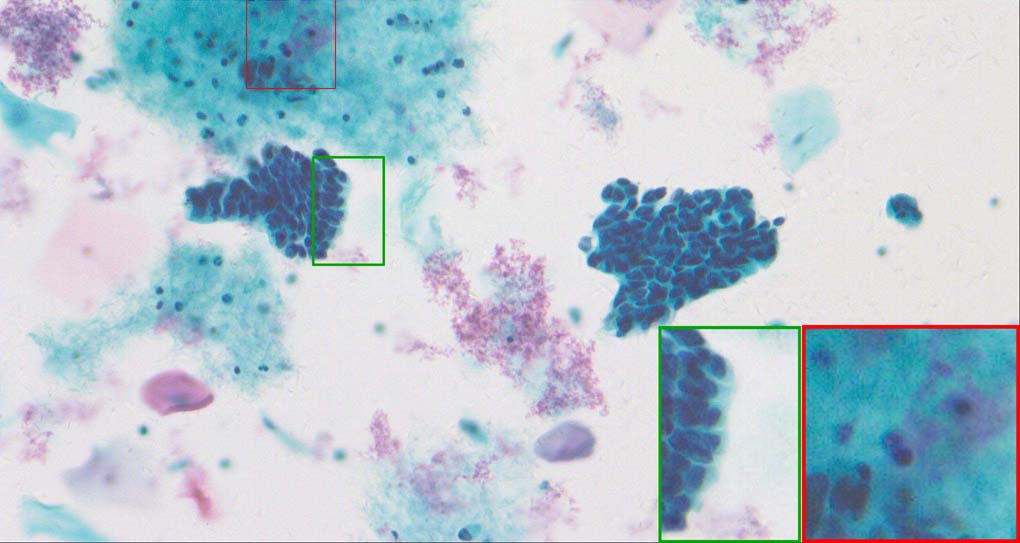}
        \caption{LP-SR}
        \label{figM1:LPSR}
    \end{subfigure}
    \begin{subfigure}[b]{0.18\textwidth}
        \includegraphics[width=\textwidth]{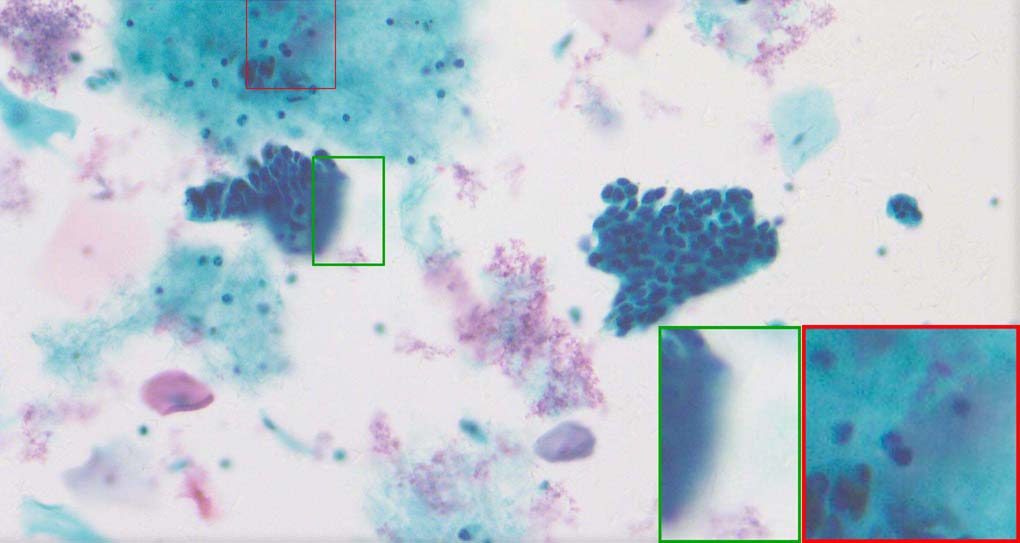}
        \caption{MWGF}
        \label{figM1:MWGF}
    \end{subfigure}
    \begin{subfigure}[b]{0.18\textwidth}
        \includegraphics[width=\textwidth]{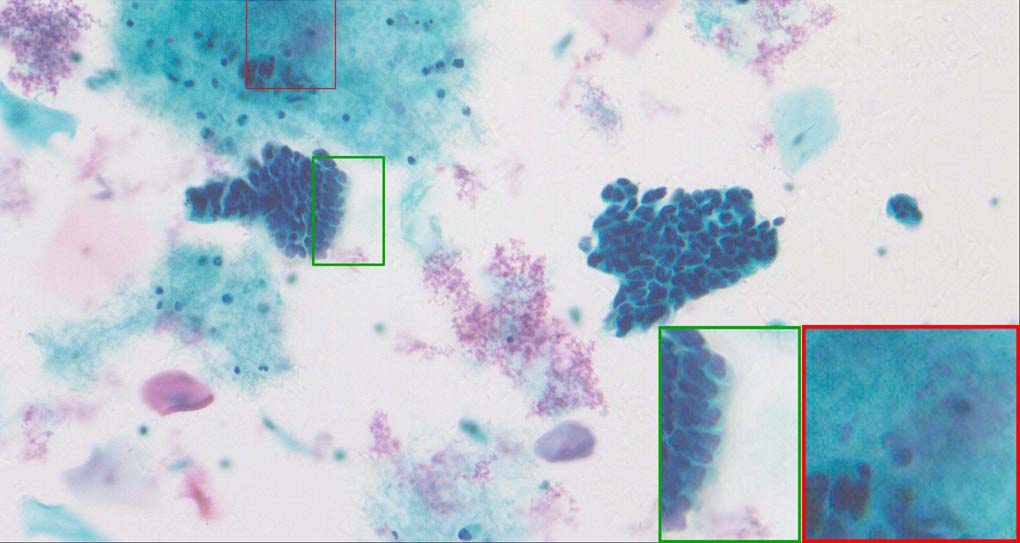}
        \caption{NSCT}
        \label{figM1:NSCT}
    \end{subfigure}
    \begin{subfigure}[b]{0.18\textwidth}
        \includegraphics[width=\textwidth]{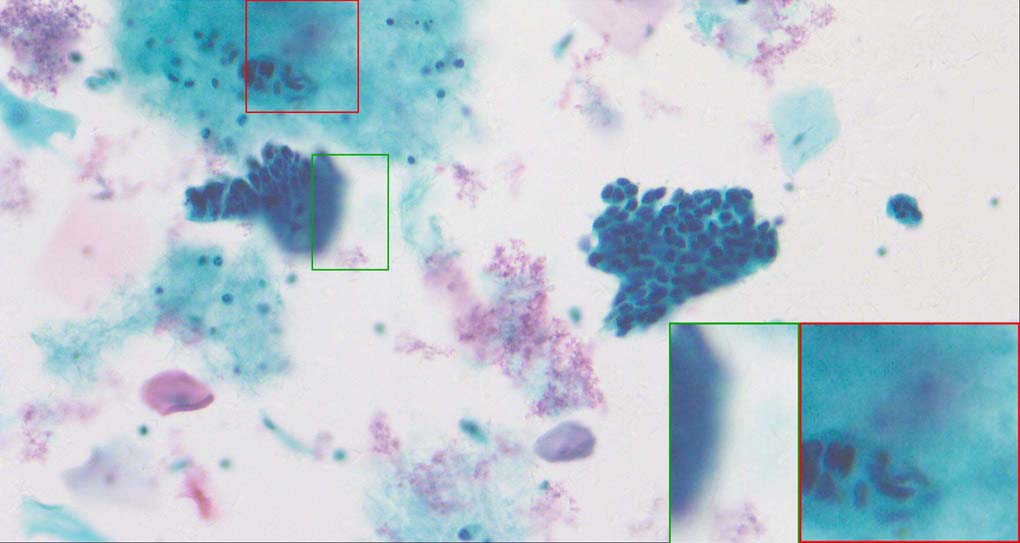}
        \caption{BF}
        \label{figM1:BF}
    \end{subfigure}
    \begin{subfigure}[b]{0.18\textwidth}
        \includegraphics[width=\textwidth]{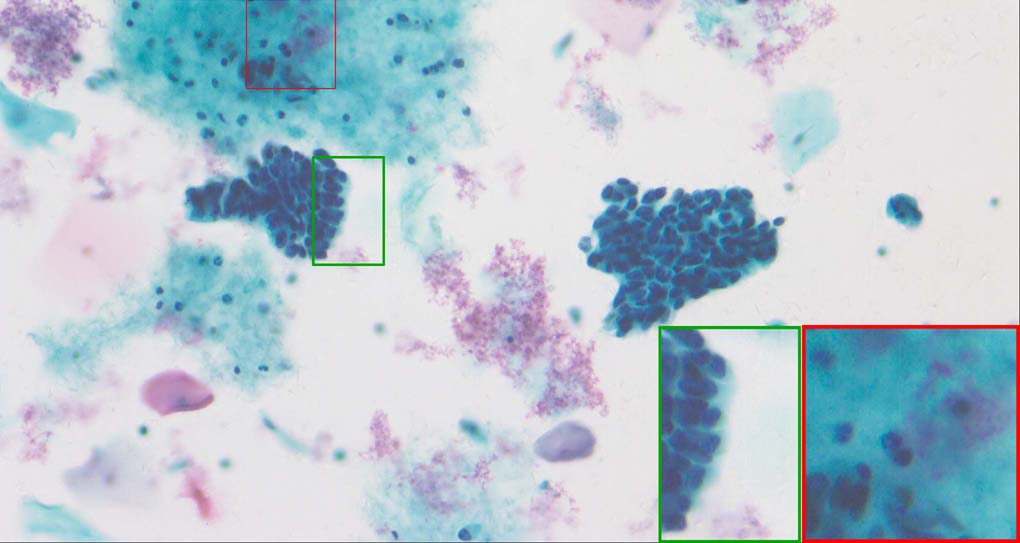}
        \caption{Proposed}
        \label{figM1:ours}
    \end{subfigure}
    \caption{Fusion results of the first group multi-focus cell images in Figure \ref{fig:cell}.}\label{CELL1_results}
\end{figure}
\begin{figure}[ht]
    \centering
    \begin{subfigure}[b]{0.18\textwidth}
        \includegraphics[width=\textwidth]{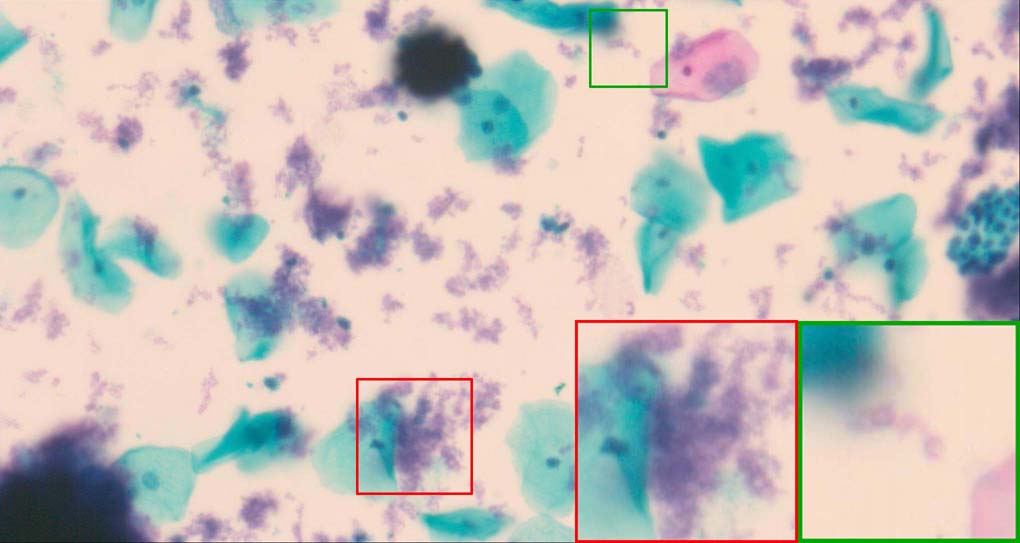}
        \caption{DSIFT}
        \label{figM2:DSIFT}
    \end{subfigure}
    \begin{subfigure}[b]{0.18\textwidth}
        \includegraphics[width=\textwidth]{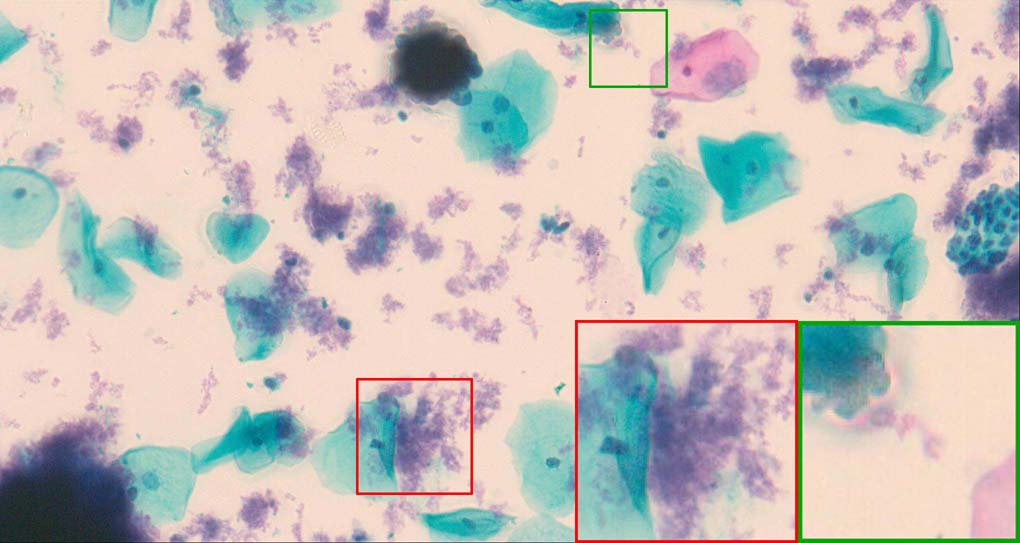}
        \caption{DTCWT}
        \label{figM2:DTCWT}
    \end{subfigure}
    \begin{subfigure}[b]{0.18\textwidth}
        \includegraphics[width=\textwidth]{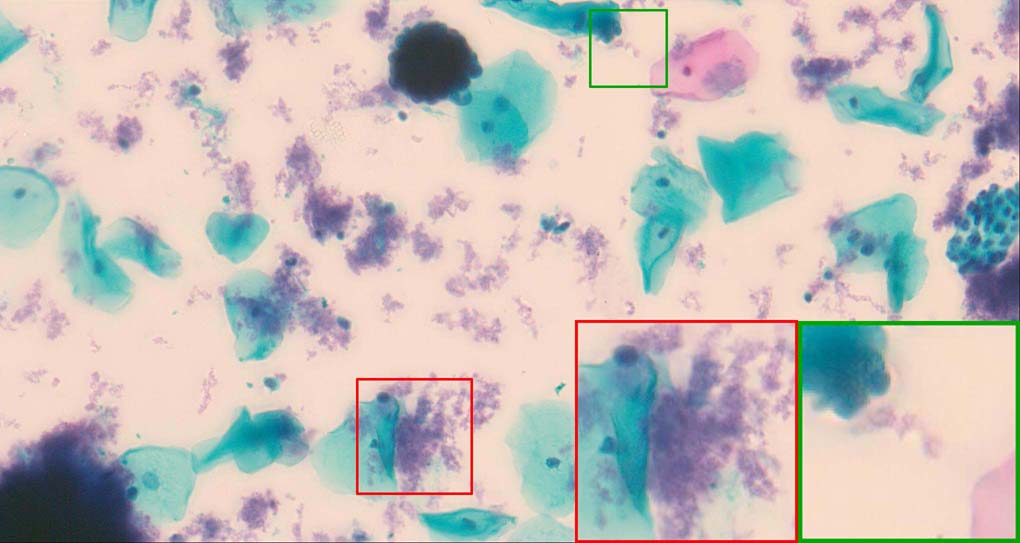}
        \caption{GFF}
        \label{figM2:gff}
    \end{subfigure}
    \begin{subfigure}[b]{0.18\textwidth}
        \includegraphics[width=\textwidth]{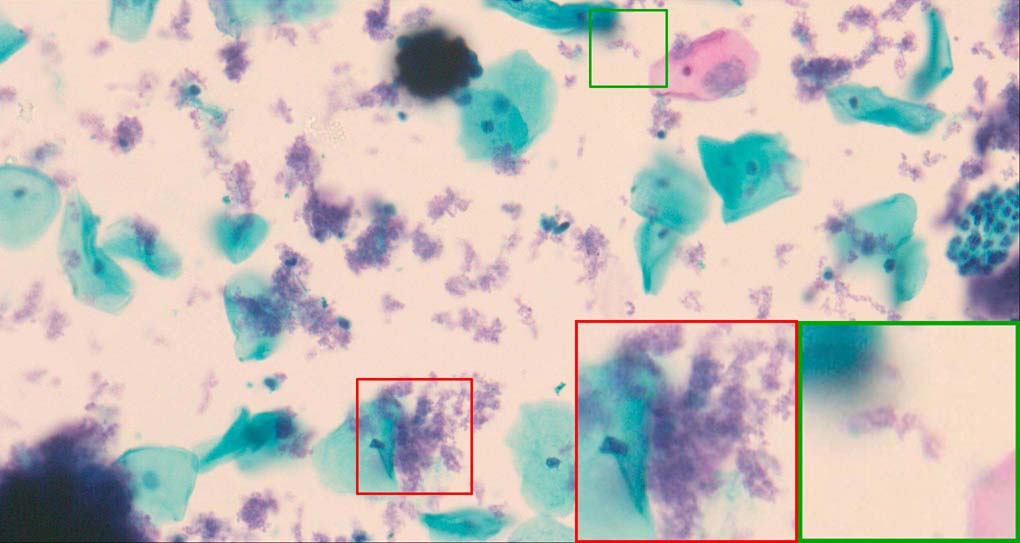}
        \caption{IM}
        \label{figM2:im}
    \end{subfigure}
    \begin{subfigure}[b]{0.18\textwidth}
        \includegraphics[width=\textwidth]{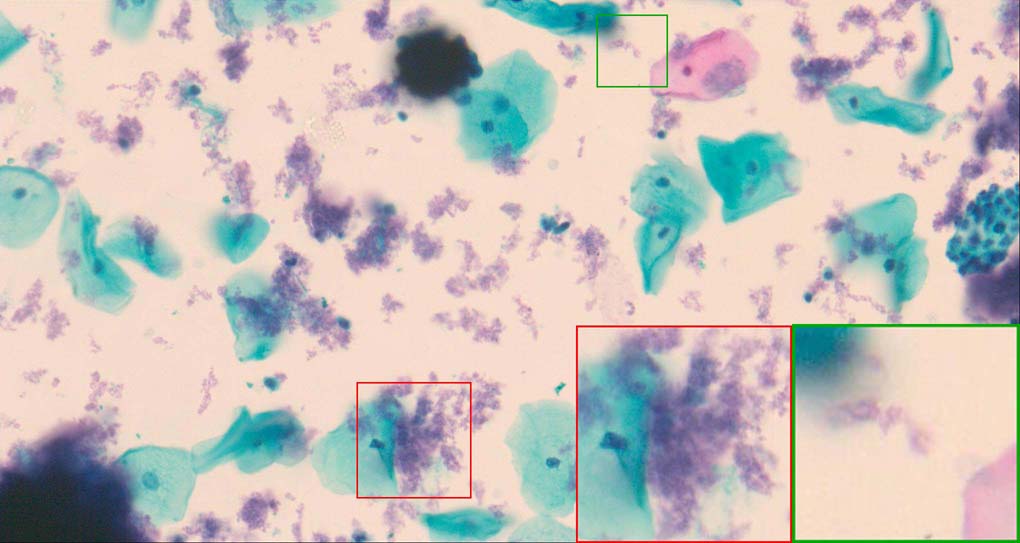}
        \caption{CNN}
        \label{figM2:CNN}
    \end{subfigure}\\
    \begin{subfigure}[b]{0.18\textwidth}
        \includegraphics[width=\textwidth]{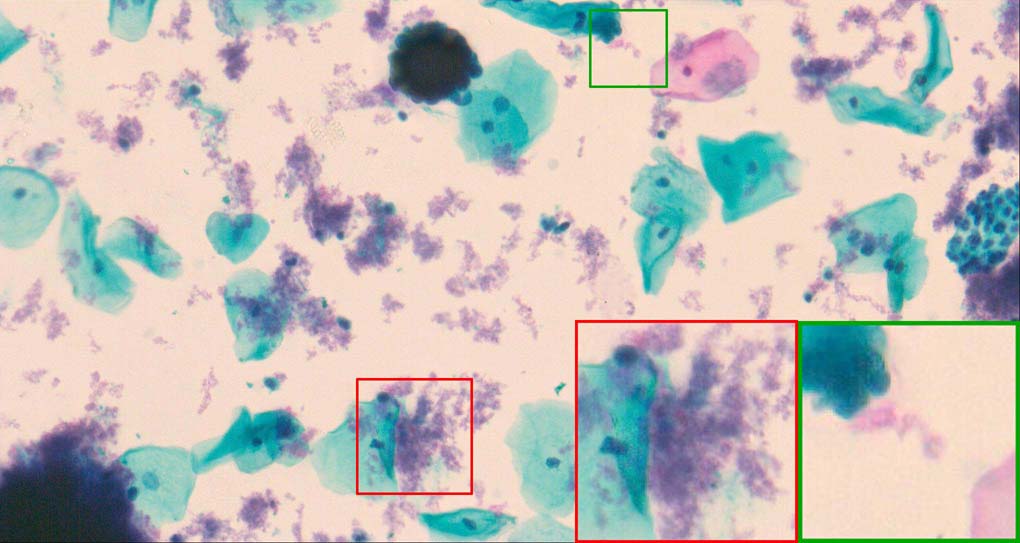}
        \caption{LP-SR}
        \label{figM2:LPSR}
    \end{subfigure}
    \begin{subfigure}[b]{0.18\textwidth}
        \includegraphics[width=\textwidth]{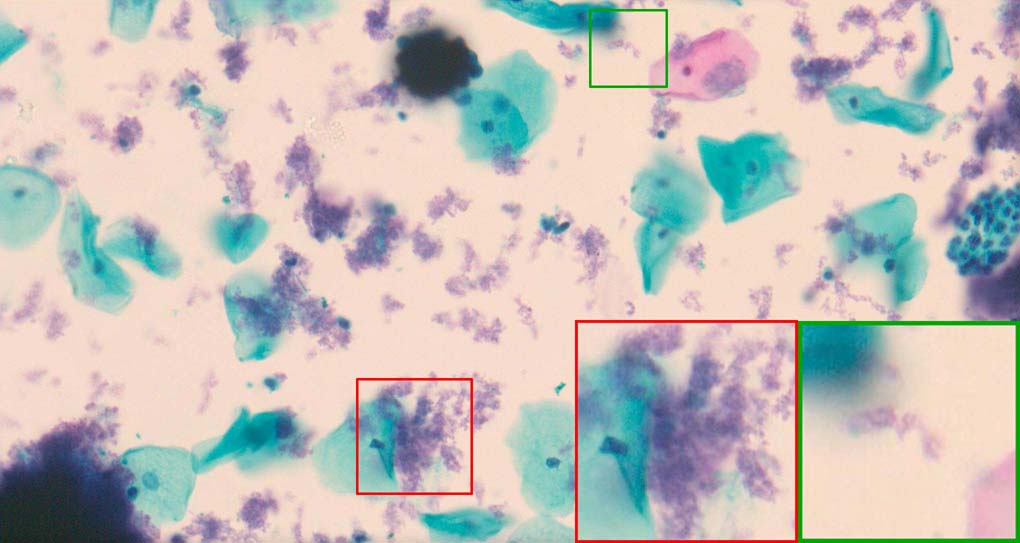}
        \caption{MWGF}
        \label{figM2:MWGF}
    \end{subfigure}
    \begin{subfigure}[b]{0.18\textwidth}
        \includegraphics[width=\textwidth]{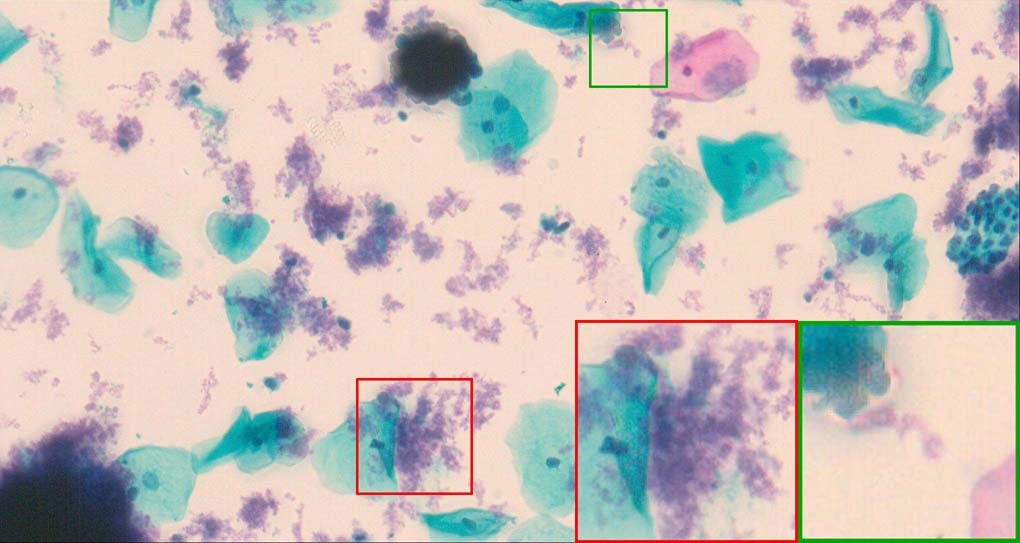}
        \caption{NSCT}
        \label{figM2:NSCT}
    \end{subfigure}
    \begin{subfigure}[b]{0.18\textwidth}
        \includegraphics[width=\textwidth]{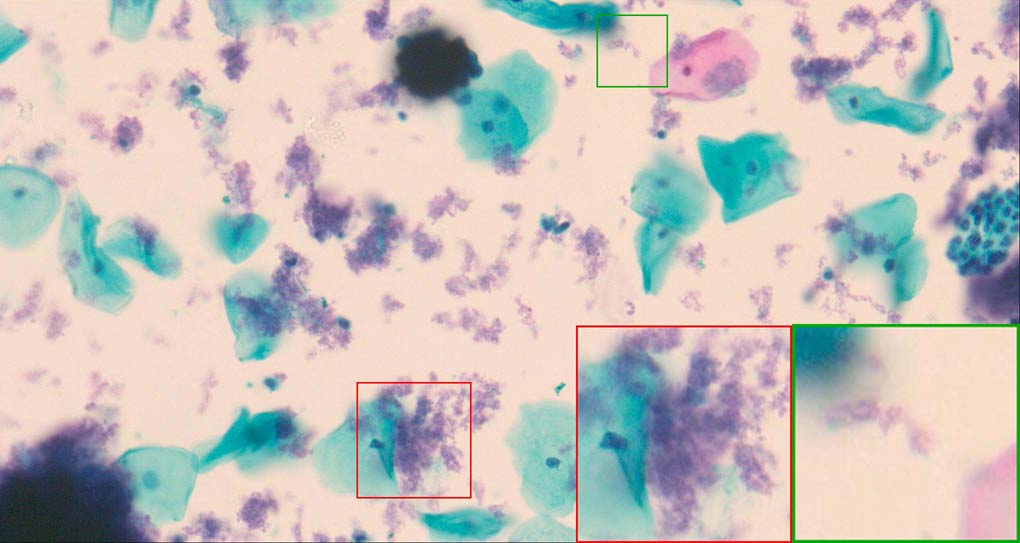}
        \caption{BF}
        \label{figM2:BF}
    \end{subfigure}
    \begin{subfigure}[b]{0.18\textwidth}
        \includegraphics[width=\textwidth]{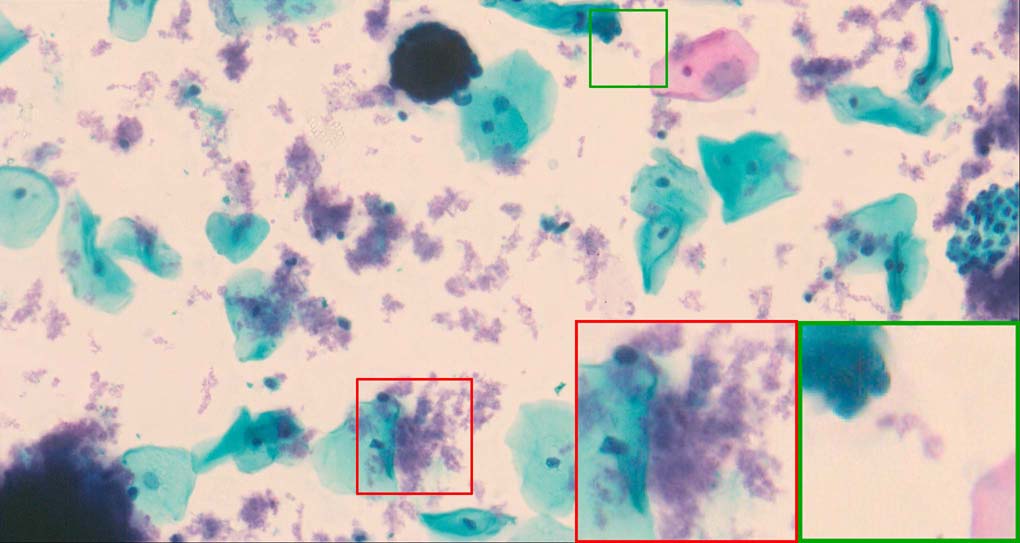}
        \caption{Proposed}
        \label{figM2:ours}
    \end{subfigure}
    \caption{Fusion results of the second group multi-focus cell images in Figure  \ref{fig:cell}.}\label{CELL2_results}
\end{figure}
\begin{figure}[ht]
    \centering
    \begin{subfigure}[b]{0.18\textwidth}
        \includegraphics[width=\textwidth]{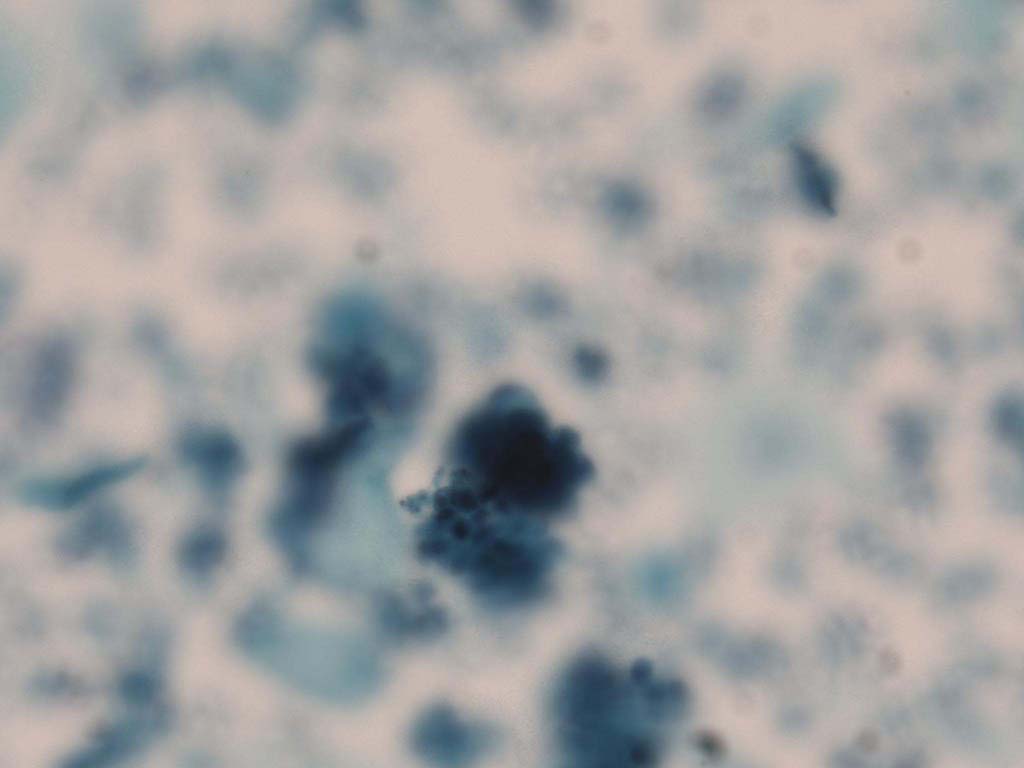}
        \caption{DSIFT}
        \label{figM3:DSIFT}
    \end{subfigure}
    \begin{subfigure}[b]{0.18\textwidth}
        \includegraphics[width=\textwidth]{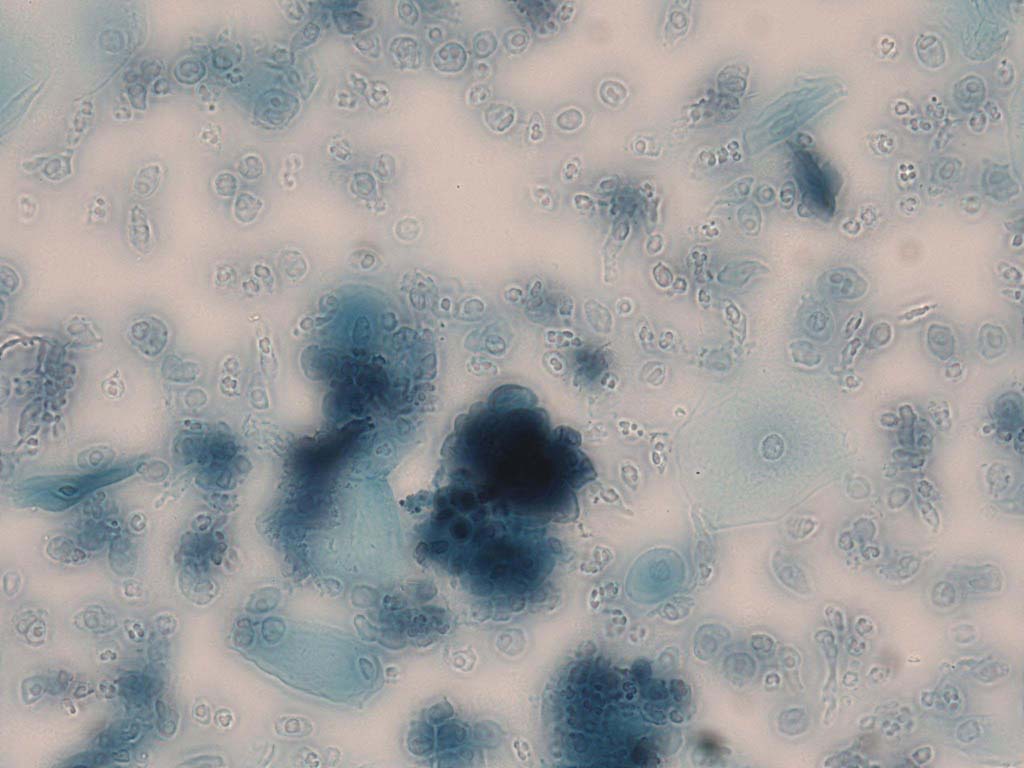}
        \caption{DTCWT}
        \label{figM3:DTCWT}
    \end{subfigure}
    \begin{subfigure}[b]{0.18\textwidth}
        \includegraphics[width=\textwidth]{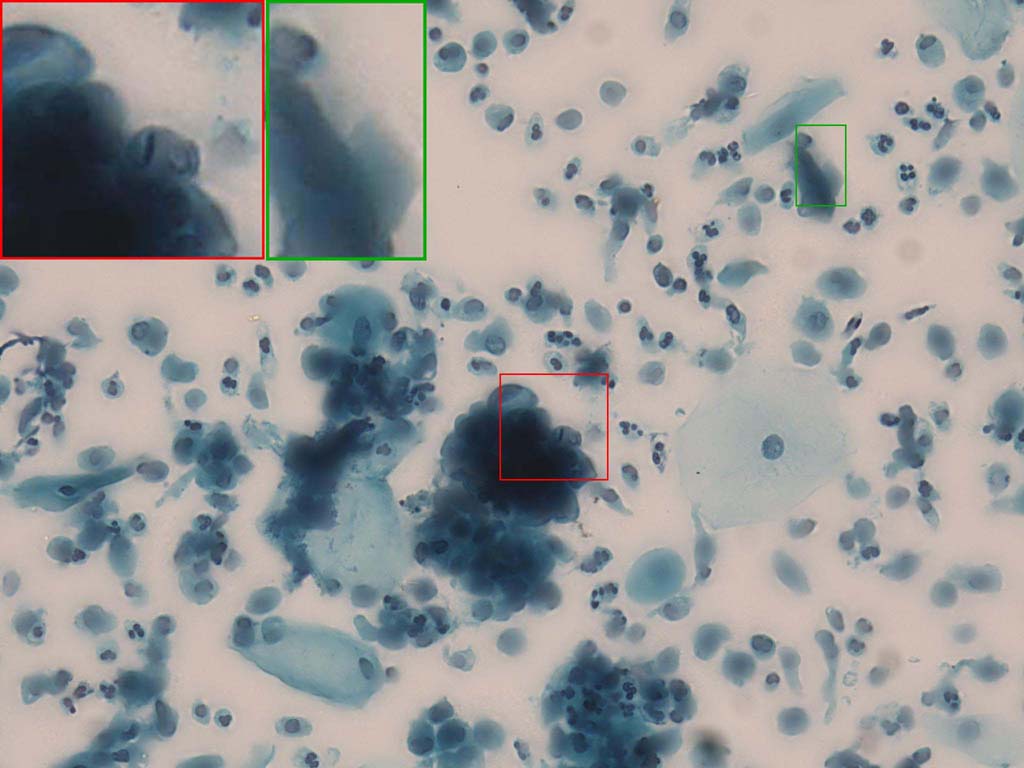}
        \caption{GFF}
        \label{figM3:gff}
    \end{subfigure}
    \begin{subfigure}[b]{0.18\textwidth}
        \includegraphics[width=\textwidth]{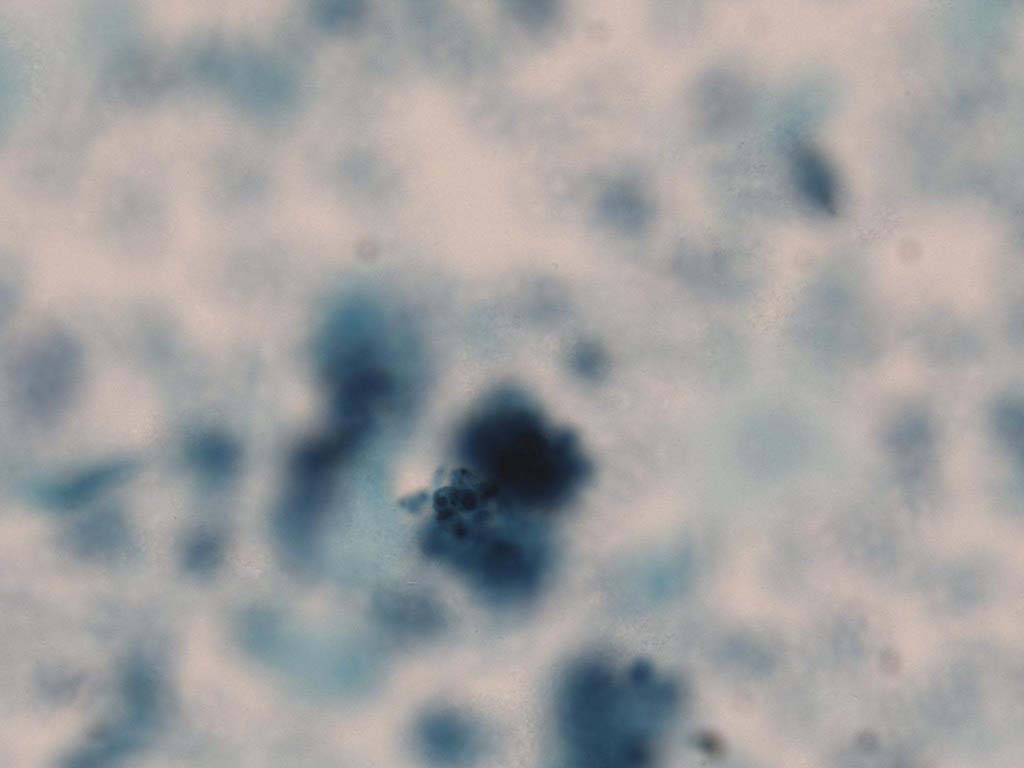}
        \caption{IM}
        \label{figM3:im}
    \end{subfigure}
    \begin{subfigure}[b]{0.18\textwidth}
        \includegraphics[width=\textwidth]{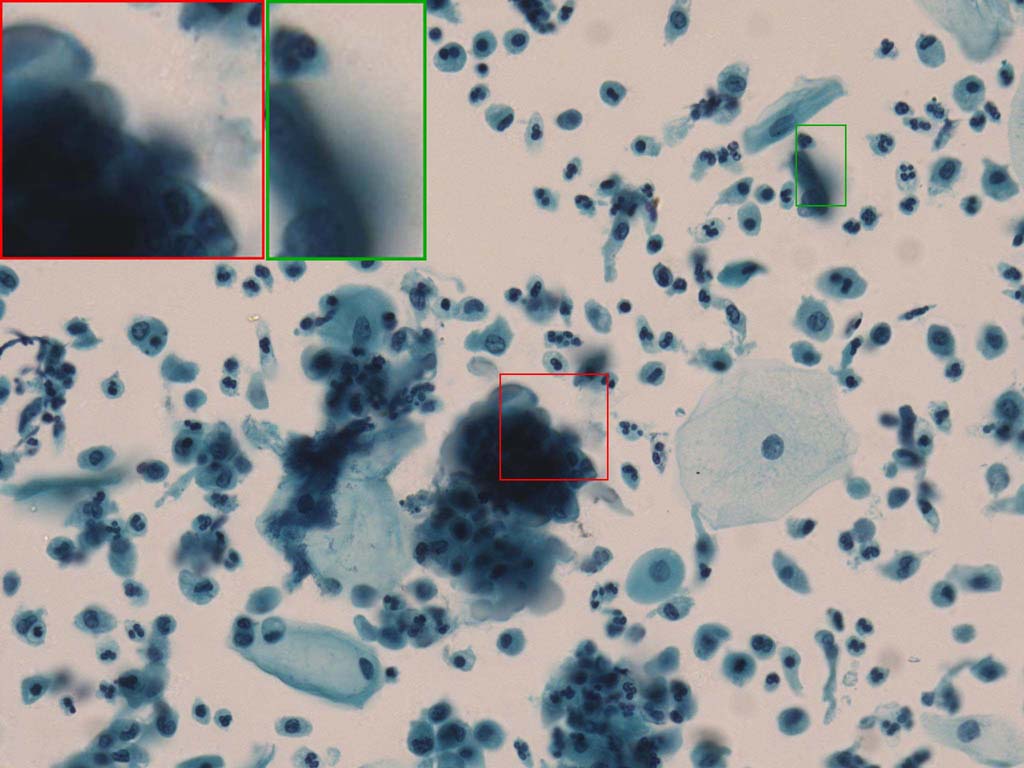}
        \caption{CNN}
        \label{figM3:CNN}
    \end{subfigure} \\
    \begin{subfigure}[b]{0.18\textwidth}
        \includegraphics[width=\textwidth]{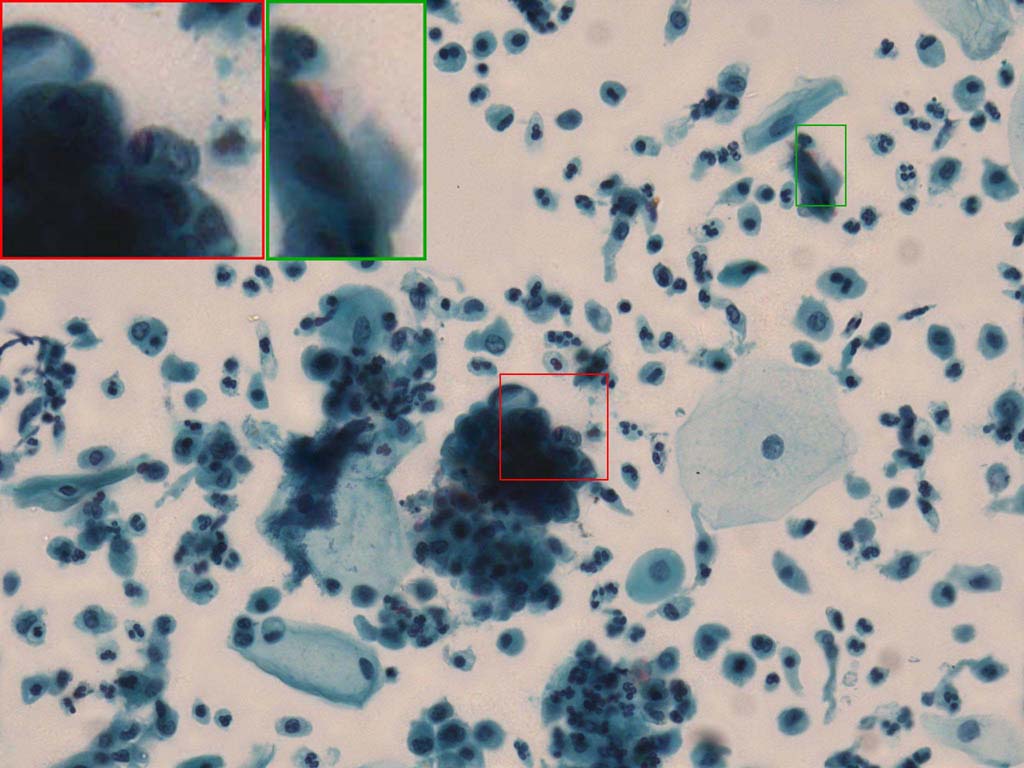}
        \caption{LP-SR}
        \label{figM3:LPSR}
    \end{subfigure}
    \begin{subfigure}[b]{0.18\textwidth}
        \includegraphics[width=\textwidth]{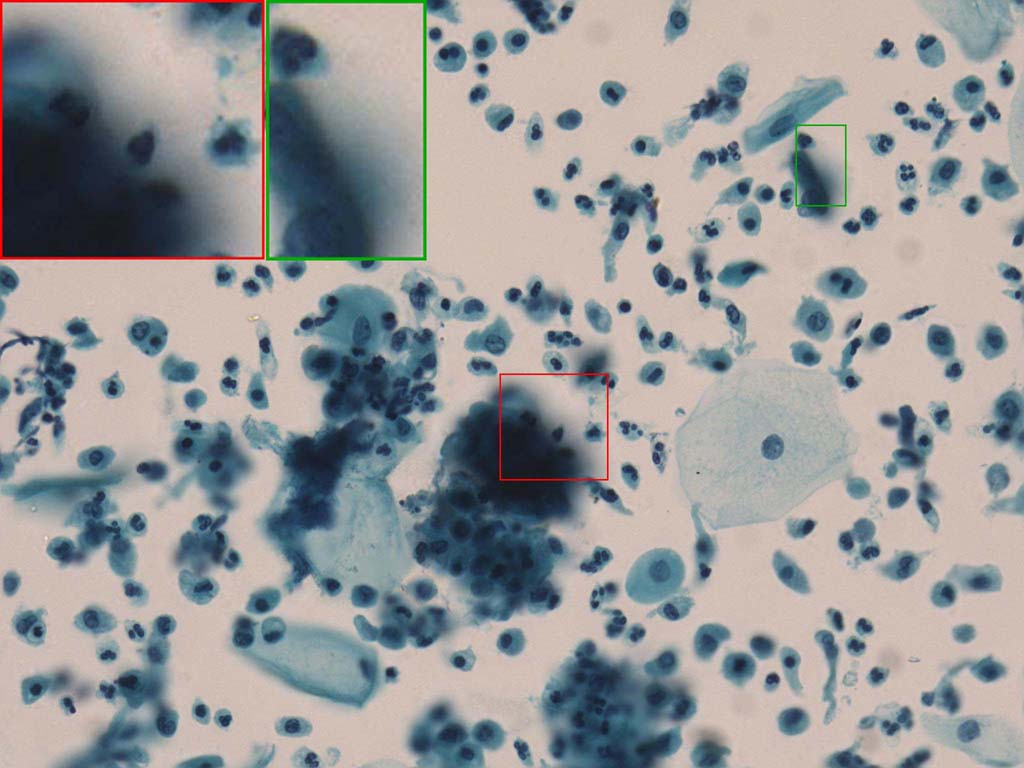}
        \caption{MWGF}
        \label{figM3:MWGF}
    \end{subfigure}
    \begin{subfigure}[b]{0.18\textwidth}
        \includegraphics[width=\textwidth]{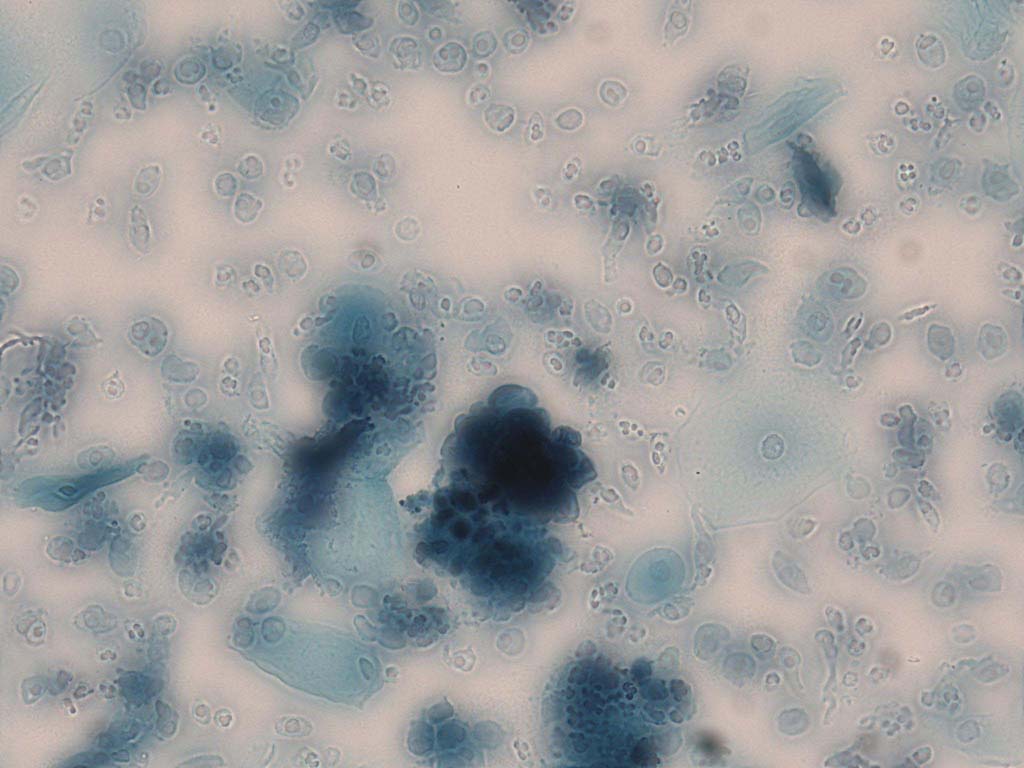}
        \caption{NSCT}
        \label{figM3:NSCT}
    \end{subfigure}
    \begin{subfigure}[b]{0.18\textwidth}
        \includegraphics[width=\textwidth]{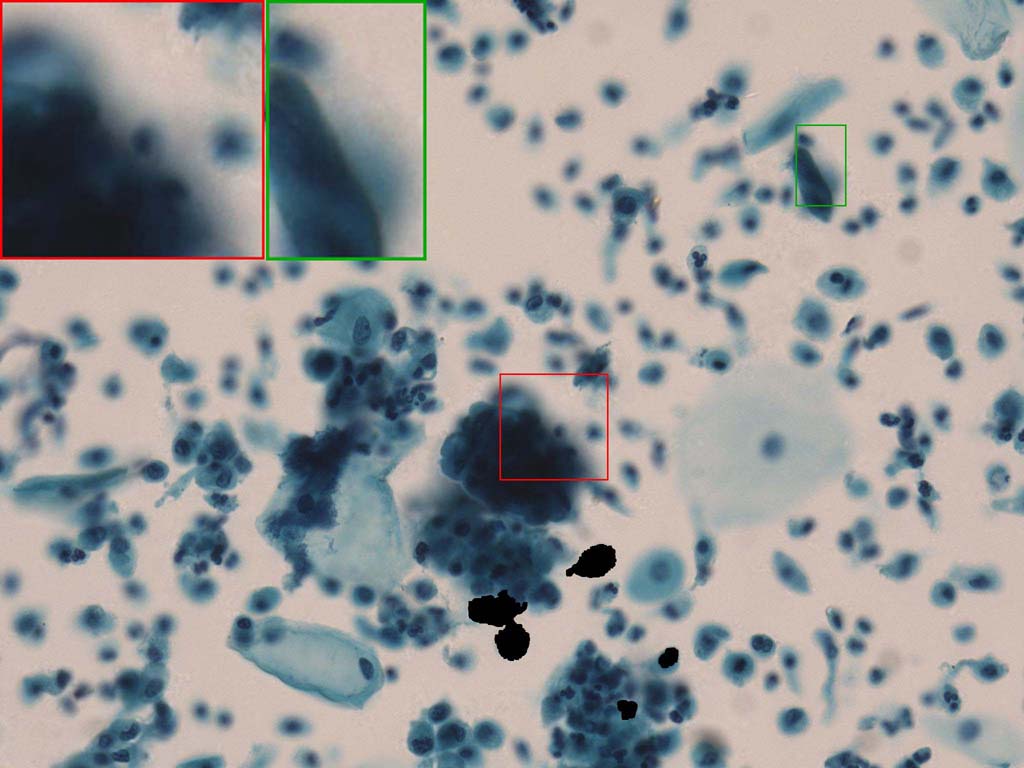}
        \caption{BF}
        \label{figM3:BF}
    \end{subfigure}
    \begin{subfigure}[b]{0.18\textwidth}
        \includegraphics[width=\textwidth]{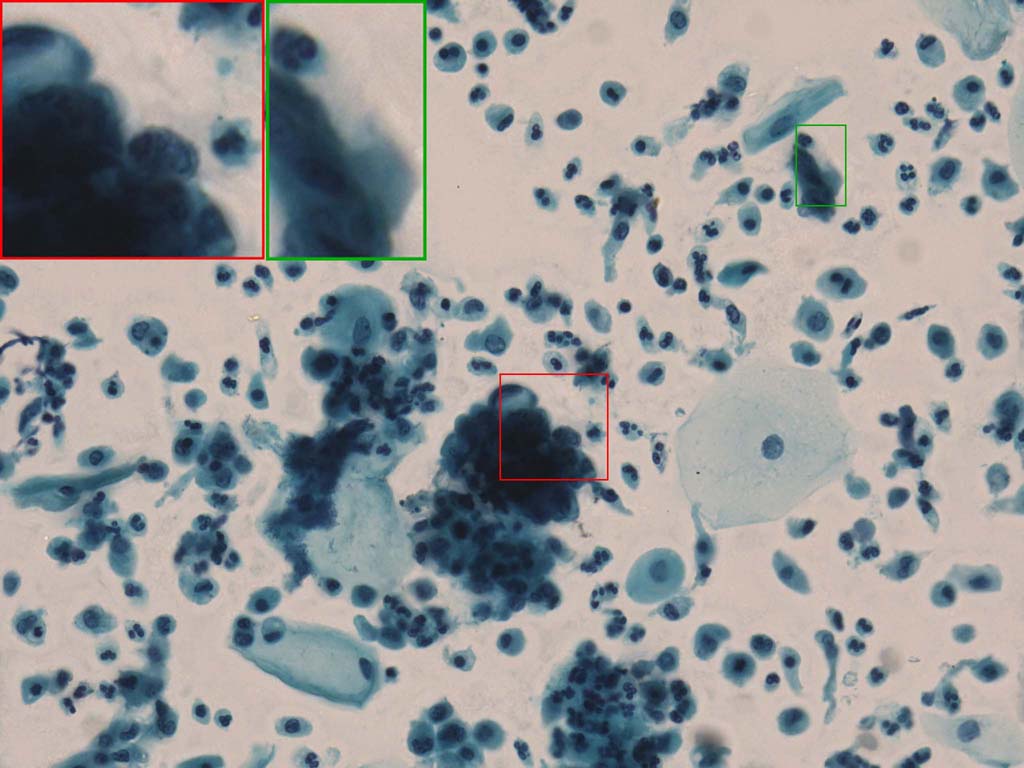}
        \caption{Proposed}
        \label{figM3:ours}
    \end{subfigure}
    \caption{Fusion results of the third group multi-focus cell images in Figure  \ref{fig:cell}.}\label{CELL3_results}
\end{figure}

Figure \ref{CELL1_results}, Figure \ref{CELL2_results} and Figure \ref{CELL3_results} show the comparative fused results of the multi-focus cell images shown in Figure \ref{fig:cell}. For clarity, we also present a closeup view in the right-bottom of each sub-picture in Figure \ref{CELL1_results} and Figure \ref{CELL2_results}. As shown in the close-up views of Figure \ref{CELL1_results}, the fused images based on DSIFT, IM, MWGF and BF methods are extremely blurred in the boundary and fail to keep the details of cell nucleus. Furthermore, the DTCWT and NSCT based methods produce halo artifacts in the fused images, while GFF and CNN based methods fail to preserve the small cell nucleus. LP-SR based method nearly works fine which keeps the most of the details of the small size cells, but the integrity of the clustered large size cells is damaged. Fortunately, in our proposed method, the integrity of the clustered large size cells is preserved and most of the isolated small size cells are maintained from the original images, which demonstrates the best visual quality.

Similarly, as shown in the close-up views of Figure \ref{CELL2_results}, the fused images from DSIFT, IM, MWGF and BF are blurred and lose some nucleus details, while the results from DTCWT, GFF, CNN and NSCT produce halo artifacts. LP-SR based method can keep details well but also produces halo artifacts and other noise. Our method can preserve the focused areas of different source images well without introducing any artifacts. For the example illustrated in Figure \ref{CELL3_results}, the fused images generated by DSIFT, DTCWT, IM and NSCT all fail to preserve the focused areas of different source images and result in extremely blurred images. The GFF, CNN, MWGF and BF based method introduces a lot of color distortion of the nucleus regions and the obvious halo artifact. The result of LP-SR based method is close to the one of our method but introduces some odd color distortion. Again, our method produces fused image which can preserve the focused areas of different source images well without introducing any artifacts.

\begin{table}[ht]
 \begin{tabular}{c|c|cccccccccc}
\hline
Source &      Index &      DSIFT &      DTCWT &        GFF &         IM &CNN&      LP-SR &       MWGF &       NSCT &BF&   Proposed \\
\hline
\multirow{6}{1.5cm}{Multi-modal dataset}   &        MI & {\bf 1.4832 } &    1.0889  &    1.2421  &    1.3495  &0.8067&    1.2566  &    1.3601  &    1.1233  &0.7921&    1.4632  \\

\multicolumn{ 1}{c|}{} &       SSIM &    0.6416  &    0.6104  &    0.6489  &    0.6460&{\bf 0.6654}  &    0.6443  &    0.6512  &    0.6342  & 0.6570& 0.6646  \\

\multicolumn{ 1}{c|}{} &      $\mathrm{Q}_{\mathrm{AB/F}}$ &  0.6017 &    0.5037  &    0.5600  &    0.5354&0.6183  &    0.5610  &    0.5983  &    0.5390 &{\bf 0.6224} &    0.5850  \\

\multicolumn{ 1}{c|}{} &         QI &    0.5109  &    0.4180  &    0.4807  &    0.5134 & {\bf 0.5962}  &    0.4887  &    0.5236  &    0.4428 &0.5931 & 0.5471  \\

\multicolumn{ 1}{c|}{} &        FMI &    0.8639  &    0.8514  &    0.8574  &    0.8560 &0.8718  &    0.8569  &  0.8680  &    0.8514  &{\bf 0.8748}&    0.8648  \\

\multicolumn{ 1}{c|}{} &        VIF & 0.3731 &    0.2310  &    0.2789  &    0.3000 &0.4165  &    0.2735  &    0.3561  &    0.2417  &{\bf 0.4444}&    0.3315  \\
\hline
\multirow{6}{1.5cm}{Natural multi-focus dataset} &        MI &    1.8910  &    1.8603  &    1.9176  &    1.9035&0.7881  &    1.8738  &    1.8840  &    1.8838  &0.7894& {\bf 2.1196 } \\

\multicolumn{ 1}{c|}{} &       SSIM &    0.8328  &    0.8271  &    0.8305  &    0.8247 &0.8431 &    0.8271  &    0.8219  &    0.8363  &0.8203& {\bf 0.8537 } \\

\multicolumn{ 1}{c|}{} &      $\mathrm{Q}_{\mathrm{AB/F}}$ &    0.6203  &    0.6196  &    0.6256  &    0.6226&{\bf 0.6774}  &    0.6206  &    0.6210  &    0.6233  &0.6660&  0.6749  \\

\multicolumn{ 1}{c|}{} &         QI &    0.6351  &    0.6182  &    0.6236  &    0.6175&0.6792  &    0.6239  &    0.6152  &    0.6381  &0.6634& {\bf 0.6891 } \\

\multicolumn{ 1}{c|}{} &        FMI &    0.8597  &    0.8612  &    0.8616  &    0.8612 &{\bf 0.8686} &    0.8615  &    0.8619  &    0.8612  &0.8583&  0.8663  \\

\multicolumn{ 1}{c|}{} &        VIF &    0.4871  &    0.4820  &    0.4953  &    0.4905&{\bf 0.5756}  &    0.4827  &    0.4948  &    0.4857  &0.5584&  0.5457  \\
\hline
 \multirow{6}{1.5cm}{Multi-focus cell dataset}&        MI & {\bf 1.3387 } &    1.1634  &    1.3042  &    1.2152&0.6893  &    1.0834  &    1.1295  &    1.1507  &0.6801&    1.1422  \\

\multicolumn{ 1}{c|}{} &       SSIM & {\bf 0.6923 } &    0.6568  &    0.6839  &    0.6552 &0.6279 &    0.6436  &    0.6481  &    0.6566  &0.6407&    0.6422  \\

\multicolumn{ 1}{c|}{} &      $\mathrm{Q}_{\mathrm{AB/F}}$ &    0.1870  &    0.2120  &    0.2100  &    0.1950 &0.2051 &    0.2331  &    0.1886  &    0.2228  &0.2102& {\bf 0.2468 } \\

\multicolumn{ 1}{c|}{} &         QI &    0.1915  &    0.1779  &    0.1844  &    0.1677 &0.1662 &    0.2161  &    0.1814  &    0.2115  &0.1855& {\bf 0.2474 } \\

\multicolumn{ 1}{c|}{} &        FMI &    0.7603  &    0.7570  & {\bf 0.7622 } &    0.7475&0.7456  &    0.7579  &    0.7152  &    0.7573  &0.7537&    0.7605  \\

\multicolumn{ 1}{c|}{} &        VIF &    0.2261  &    0.1854  &    0.2205  &    0.1959&0.2087  &    0.2293  &    0.2225  &    0.1967 &0.2270 & {\bf 0.2388 } \\

\hline
 \end{tabular}
  \caption{The average objective assessments of different fusion methods.}
  \label{Table}
\end{table}
\begin{table}[ht]

\centering
  \begin{tabular}{c|c c c c c c c c c |c c}
  \hline
           & DSIFT & DTCWT & IM &CNN& LP-SR & MWGF & NSCT &BF&GFF &GFF(C++)&  Proposed\\
  \hline
    Time(s)& 1679.67 & 36.53 &  72.47 &5276.09& 34.29  & 259.30 & 414.24 &453.39&10.66 &6.44& \textbf{2.08} \\
  \hline
  \end{tabular}
 \caption{The average running time of different fusion methods on color cell images with size of $2040 \times 1086$.}
  \label{Time_consumption}
\end{table}

The quantitative results of different fusion methods are shown in Table \ref{Table}. It can be seen that the proposed method yields competitive objective metrics on the natural multi-focus dataset and multi-focus
cell dataset. For the multi-modal dataset, the metrics value is not the best but is nearly close to the best performance, such as the value of MI, SSIM, FMI. We also compare the computational efficiency of each methods on the high-resolution color cell images with the size of $2040 \times 1086$. Experiments are performed on a computer equipped with a 4.20 GHz CPU and 8GB memory and all codes are available online. The average running time of different image fusion methods is compared in Table \ref{Time_consumption}. As mentioned before the method of DSIFT, DTCWT, GFF, IM, LP-SR, MWGF, NSCT and BF are all implemented in Matlab while the CNN-based and our method are based on C++, and therefore strictly speaking, the comparison is running time unfair. Here, we re-implement the GFF-based method with C++ and also include the corresponding running time in Table \ref{Time_consumption} to reveal the running efficiency of different implementation between Matlab and C++ to some extent. As shown in Table \ref{Time_consumption}, the guided filtering based methods, i.e. GFF-based and the proposed method are the most efficient methods while the CNN-based is the most time-consuming. Comparing to the original Matlab implementation, the GFF-method can be speeded up by almost 40\% with C++ implementation, but it is still much slower than our method. We attribute this to the following reasons. First, the computation burden of the DoG-based scale-invariant saliency selection step can be negligible comparing to the computation burden of activity maps refinement step based on the guided filtering. The GFF-based method needs to perform guided filtering twice (each for both base and detail layers), while our method only needs to perform filtering one time to refine the activity maps of each source image. Second, instead of using the original color image, we use the gray one as the guided image to accelerate the activity refinement step. Due to the extremely efficiency, our method can be applied for some nearly real-time applications such as digital cytopathology \cite{pantanowitz2009impact} and can be furthermore accelerated through GPU programming.

\section{Conclusion}
In this paper, based on the scale-space theory we propose a very simple yet effective multi-scale image fusion method in spatial domain. To keep both details of small size objects and the integrity information of large size objects in the fused image, we first get a robust saliency map with the \emph{scale-invariant} structure based on the DoG pyramid, which transfers the details and integrity detection of the objects into a scale-zooming intensive response. Then the activity map is constructed by \emph{non-max suppression} scheme based on the saliency maps and refined by the guided filtering to capture the spatial context. Finally, the fused image is generated by combining the activity maps and the original input images intuitively. Experimental results demonstrate that our method is efficient and can produce an all-in-focus image with a high quality, which can preserve the details and the integrity of very different size objects well. Meanwhile, due to the low-time complexity, the propose method can deal with high resolution images in a very efficient way and can be applied for the real-time application.

\section*{Acknowledgements}
This research was partially supported by the Natural Science Foundation of Hunan Province, China (No.14JJ2008) and the National Natural Science Foundation of China under Grant No. 61602522, No. 61573380, No. 61672542 and the Fundamental Research Funds of the Central Universities of Central South University under Grant No. 2018zzts577.

\bibliographystyle{plainnat}
\bibliography{reference}

\begin{thebibliography}{10}
\expandafter\ifx\csname url\endcsname\relax
  \def\url#1{\texttt{#1}}\fi
\expandafter\ifx\csname urlprefix\endcsname\relax\def\urlprefix{URL }\fi
\expandafter\ifx\csname href\endcsname\relax
  \def\href#1#2{#2} \def\path#1{#1}\fi

\bibitem{stathaki2011image}
T.~Stathaki, Image fusion: Algorithms and applications, Elsevier, 2011.

\bibitem{li2017pixel}
S.~Li, X.~Kang, L.~Fang, J.~Hu, H.~Yin, Pixel-level image fusion: A survey of
  the state of the art, Information Fusion 33 (2017) 100--112.

\bibitem{du2016overview}
J.~Du, W.~Li, K.~Lu, B.~Xiao, An overview of multi-modal medical image fusion,
  Neurocomputing 215 (2016) 3--20.

\bibitem{duan2018multifocus}
J.~Duan, L.~Chen, C.~P. Chen, Multifocus image fusion with enhanced linear
  spectral clustering and fast depth map estimation, Neurocomputing 318 (2018)
  43--54.

\bibitem{nayar2015bethesda}
R.~Nayar, D.~C. Wilbur, The Bethesda system for reporting cervical cytology:
  Definitions, criteria, and explanatory notes, Springer, 2015.

\bibitem{pantanowitz2009impact}
L.~Pantanowitz, M.~Hornish, R.~A. Goulart, The impact of digital imaging in the
  field of cytopathology, Cytojournal 6.

\bibitem{lowe2004distinctive}
D.~G. Lowe, Distinctive image features from scale-invariant keypoints,
  International Journal of Computer Vision 60~(2) (2004) 91--110.

\bibitem{he2013guided}
K.~He, J.~Sun, X.~Tang, Guided image filtering, IEEE Transactions on Pattern
  Analysis and Machine Intelligence 35~(6) (2013) 1397--1409.

\bibitem{he2018multi}
K.~He, D.~Zhou, X.~Zhang, R.~Nie, Multi-focus: Focused region finding and
  multi-scale transform for image fusion, Neurocomputing 320 (2018) 157--170.

\bibitem{liu2017structure}
X.~Liu, W.~Mei, H.~Du, Structure tensor and nonsubsampled shearlet transform
  based algorithm for {CT} and {MRI} image fusion, Neurocomputing 235 (2017)
  131--139.

\bibitem{yang2010multifocus}
B.~Yang, S.~Li, Multifocus image fusion and restoration with sparse
  representation, IEEE Transactions on Instrumentation and Measurement 59~(4)
  (2010) 884--892.

\bibitem{liu2016image}
Y.~Liu, X.~Chen, R.~K. Ward, Z.~J. Wang, Image fusion with convolutional sparse
  representation, IEEE Signal Processing Letters 23~(12) (2016) 1882--1886.

\bibitem{zhang2018robust}
Q.~Zhang, T.~Shi, F.~Wang, R.~S. Blum, J.~Han, Robust sparse representation
  based multi-focus image fusion with dictionary construction and local spatial
  consistency, Pattern Recognition 83 (2018) 299--313.

\bibitem{shahdoosti2016combining}
H.~R. Shahdoosti, H.~Ghassemian, Combining the spectral {PCA} and spatial {PCA}
  fusion methods by an optimal filter, Information Fusion 27 (2016) 150--160.

\bibitem{gangapure2015steerable}
V.~N. Gangapure, S.~Banerjee, A.~S. Chowdhury, Steerable local frequency based
  multispectral multifocus image fusion, Information Fusion 23 (2015) 99--115.

\bibitem{li2013image}
S.~Li, X.~Kang, J.~Hu, Image fusion with guided filtering, IEEE Transactions on
  Image Processing 22~(7) (2013) 2864--2875.

\bibitem{chen2018robust}
Y.~Chen, J.~Guan, W.-K. Cham, Robust multi-focus image fusion using edge model
  and multi-matting, IEEE Transactions on Image Processing 27~(3) (2018)
  1526--1541.

\bibitem{li2013image2}
S.~Li, X.~Kang, J.~Hu, B.~Yang, Image matting for fusion of multi-focus images
  in dynamic scenes, Information Fusion 14~(2) (2013) 147--162.

\bibitem{Zhang2017Boundary}
Y.~Zhang, X.~Bai, T.~Wang, Boundary finding based multi-focus image fusion
  through multi-scale morphological focus-measure, Information Fusion 35 (2017)
  81--101.

\bibitem{liu2018deep}
Y.~Liu, X.~Chen, Z.~Wang, Z.~J. Wang, R.~K. Ward, X.~Wang, Deep learning for
  pixel-level image fusion: Recent advances and future prospects, Information
  Fusion 42 (2018) 158--173.

\bibitem{Liu2017Multi}
Y.~Liu, X.~Chen, H.~Peng, Z.~Wang, Multi-focus image fusion with a deep
  convolutional neural network, Information Fusion 36 (2017) 191--207.

\bibitem{du2017image}
C.~Du, S.~Gao, Image segmentation-based multi-focus image fusion through
  multi-scale convolutional neural network, IEEE Access 5~(99) (2017)
  15750--15761.

\bibitem{zhang2009multifocus}
Q.~Zhang, B.-l. Guo, Multifocus image fusion using the nonsubsampled contourlet
  transform, Signal Processing 89~(7) (2009) 1334--1346.

\bibitem{zhou2014multi}
Z.~Zhou, S.~Li, B.~Wang, Multi-scale weighted gradient-based fusion for
  multi-focus images, Information Fusion 20 (2014) 60--72.

\bibitem{liu2015general}
Y.~Liu, S.~Liu, Z.~Wang, A general framework for image fusion based on
  multi-scale transform and sparse representation, Information Fusion 24 (2015)
  147--164.

\bibitem{shen2013cross}
R.~Shen, I.~Cheng, A.~Basu, Cross-scale coefficient selection for volumetric
  medical image fusion, IEEE Transactions on Biomedical Engineering 60~(4)
  (2013) 1069--1079.

\bibitem{zagoruyko2015learning}
S.~Zagoruyko, N.~Komodakis, Learning to compare image patches via convolutional
  neural networks, in: Proceedings of the IEEE Conference on Computer Vision
  and Pattern Recognition, 2015, pp. 4353--4361.

\bibitem{lindeberg1994scale}
T.~Lindeberg, Scale-space theory: A basic tool for analyzing structures at
  different scales, Journal of Applied Statistics 21~(1-2) (1994) 225--270.

\bibitem{mikolajczyk2004scale}
K.~Mikolajczyk, C.~Schmid, Scale \& affine invariant interest point detectors,
  International Journal of Computer Vision 60~(1) (2004) 63--86.

\bibitem{mikolajczyk2002detection}
K.~Mikolajczyk, Detection of local features invariant to affines
  transformations, Ph.D. thesis, Institut National Polytechnique de Grenoble
  (2002).

\bibitem{kolmogorov2004energy}
V.~Kolmogorov, R.~Zabin, What energy functions can be minimized via graph
  cuts?, IEEE Transactions on Pattern Analysis and Machine Intelligence 26~(2)
  (2004) 147--159.

\bibitem{petschnigg2004digital}
G.~Petschnigg, R.~Szeliski, M.~Agrawala, M.~Cohen, H.~Hoppe, K.~Toyama, Digital
  photography with flash and no-flash image pairs, in: ACM Transactions on
  Graphics (TOG), Vol.~23, ACM, 2004, pp. 664--672.

\bibitem{liu2015multi}
Y.~Liu, S.~Liu, Z.~Wang, Multi-focus image fusion with dense {SIFT},
  Information Fusion 23 (2015) 139--155.

\bibitem{jia2014caffe}
Y.~Jia, E.~Shelhamer, J.~Donahue, S.~Karayev, J.~Long, R.~Girshick,
  S.~Guadarrama, T.~Darrell, Caffe: Convolutional architecture for fast feature
  embedding, in: Proceedings of the 22nd ACM International Conference on
  Multimedia, ACM, 2014, pp. 675--678.

\bibitem{Qu2002Information}
G.~Qu, D.~Zhang, P.~Yan, Information measure for performance of image fusion,
  Electronics Letters 38~(7) (2002) 313--315.

\bibitem{wang2004image}
Z.~Wang, A.~C. Bovik, H.~R. Sheikh, E.~P. Simoncelli, Image quality assessment:
  From error visibility to structural similarity, IEEE Transactions on Image
  Processing 13~(4) (2004) 600--612.

\bibitem{wang2002universal}
Z.~Wang, A.~C. Bovik, A universal image quality index, IEEE Signal Processing
  Letters 9~(3) (2002) 81--84.

\bibitem{xydeas2000objective}
C.~Xydeas, V.~Petrovic, Objective image fusion performance measure, Electronics
  Letters 36~(4) (2000) 308--309.

\bibitem{haghighat2011non}
M.~B.~A. Haghighat, A.~Aghagolzadeh, H.~Seyedarabi, A non-reference image
  fusion metric based on mutual information of image features, Computers \&
  Electrical Engineering 37~(5) (2011) 744--756.

\bibitem{sheikh2006image}
H.~R. Sheikh, A.~C. Bovik, Image information and visual quality, IEEE
  Transactions on Image Processing 15~(2) (2006) 430--444.

\end{thebibliography}
\end{document}